\documentclass{article} 
\usepackage{geometry}
\usepackage{url}

\usepackage{graphicx} 
\usepackage{pgf,tikz}
\usetikzlibrary{arrows,automata,fit,positioning}

\usepackage[latin9]{inputenc}
\usepackage{amsmath}
\usepackage{amssymb}
\usepackage{amsthm}
\newtheorem{theorem}{Theorem}
\newtheorem{lemma}{Lemma}
\usepackage{sidecap}  
\usepackage{subfig}
\usepackage{bm,amsbsy} 

\usepackage[ plainpages = true, pdfpagelabels, 
               pdfpagelayout = useoutlines,
               bookmarks,
               bookmarksopen = true,
               bookmarksnumbered = true,
               breaklinks = true,
               linktocpage,
               pagebackref,
               colorlinks = true,
               linkcolor = blue,
               urlcolor  = blue,
               citecolor = blue,
               anchorcolor = green,
               hyperindex = true,
               hyperfigures
               ]{hyperref}

\newcommand{\diag}{\mathop{\mathrm{diag}}}
\newcommand{\trace}{\mathop{\mathrm{tr}}}

\newcommand{\Nrm}{\mathcal{N}}
\newcommand{\Dat}{\mathcal{D}}

\newcommand{\vx}{\mathbf{x}}

\newcommand{\vv}{\mathbf{v}}

\newcommand{\vu}{\mathbf{u}}
\newcommand{\vy}{\mathbf{y}}

\newcommand{\vones}{\mathbf{1}}
\newcommand{\Em}{\mathbb{E}}
 
\newcommand{\trp}{{^\top}} 
\renewcommand{\b}{\mathbf} 
\newcommand{\bC}{\mathbf{C}} 

\newcommand{\vm}{\mathbf{m}}

\newcommand{\ve}{\mathbf{e}} 

\newcommand{\vb}{\mathbf{b}}

\newcommand{\vc}{\mathbf{c}}  
\newcommand{\vw}{\mathbf{w}} 
 
\newcommand{\vq}{\mathbf{q}} 

\newcommand{\vmu}{\mathbf{\ensuremath{\bm{\mu}}}}

\newcommand{\vtheta}{\mathbf{\ensuremath{\bm{\theta}}}}

\newcommand{\graph}{\mathcal{G}}
\newcommand{\edges}{\mathcal{E}_{\mathcal{G}}}

\DeclareMathOperator*{\argmax}{arg\,max}

\renewcommand{\eqref}[1]{Eq.~(\ref{eq:#1})}
\newcommand{\figref}[1]{Fig.~\ref{fig:#1}}
\newcommand{\secref}[1]{section~\ref{sec:#1}}

\usepackage{color}
\definecolor{darkred}{rgb}{0.6, 0, 0}
\definecolor{gray}{RGB}{0.5,0.5,0.5}

\title{Bayesian Manifold Learning: \\
The Locally Linear Latent Variable Model }

\author{Mijung Park, \, Wittawat Jitkrittum,  \,  Ahmad Qamar\thanks{Current
affiliation: Thread Genius}, \\ Zolt{\'a}n Szab{\'o},  \,
Lars Buesing\thanks{Current affiliation: Google DeepMind}, \,\, Maneesh Sahani \\ [3mm]
Gatsby Computational Neuroscience Unit\\ 
University College London\\ [3mm]
\texttt{ \{mijung, wittawat, zoltan.szabo\}@gatsby.ucl.ac.uk} \\
\texttt{ atqamar@gmail.com, lbuesing@google.com, maneesh@gatsby.ucl.ac.uk} 
}

\begin{document}

\maketitle

\begin{abstract}
We introduce the \textit{Locally Linear Latent Variable Model}
(LL-LVM), a probabilistic model for non-linear manifold discovery that
describes a joint distribution over observations, their manifold
coordinates and 
locally linear maps 
conditioned on a set of
neighbourhood relationships.  
The model allows  straightforward variational optimisation of the
posterior distribution on coordinates and locally linear maps from the
latent space to the observation space given the data.  Thus, the
LL-LVM encapsulates the local-geometry preserving intuitions that
underlie non-probabilistic methods such as locally linear embedding
(LLE).  Its probabilistic semantics make it easy to evaluate the
quality of hypothesised neighbourhood relationships, select the
intrinsic dimensionality of the manifold, construct out-of-sample
extensions and to combine the manifold model with additional
probabilistic models that capture the structure of coordinates within
the manifold.
\end{abstract}


\section{Introduction}
\label{sec:intro}


Many high-dimensional datasets comprise points derived from a smooth,
lower-dimensional manifold embedded within the high-dimensional space
of measurements and possibly corrupted by noise.  For instance, 
biological or medical imaging data might reflect the
interplay of a small number of latent processes that all affect
measurements non-linearly.  Linear multivariate analyses such as
principal component analysis (PCA) or multidimensional scaling (MDS)
have long been used to estimate such underlying processes, but cannot
always reveal low-dimensional structure when the mapping is non-linear
(or, equivalently, the manifold is curved).  Thus, there has been
substantial recent interest in algorithms to identify non-linear
manifolds in data.


Many more-or-less heuristic methods for non-linear manifold discovery
are based on the idea of preserving the geometric properties of local
neighbourhoods within the data, while embedding, unfolding or
otherwise transforming the data to occupy fewer dimensions.  Thus,
algorithms such as locally-linear embedding (LLE) and Laplacian
eigenmap attempt to preserve local linear relationships or to
minimise the distortion of local derivatives 
\cite{roweis2000reduction, Belkin02laplacianeigenmaps}.  Others, like
Isometric feature mapping (Isomap) or maximum variance unfolding (MVU)
preserve local distances, estimating global manifold properties by
continuation across neighbourhoods before embedding to lower
dimensions by classical methods such as PCA or MDS
\cite{TenenbaumEtAl2000}. 
While generally hewing to
this same intuitive path, the range of available algorithms has grown
very substantially in recent years
\cite{Maaten08dimensionalityreduction:, cayton2005algorithms}.

However, these approaches do not define distributions over the data or
over the manifold properties.  Thus, 
they provide no measures of uncertainty on manifold
structure or on the low-dimensional locations of the embedded points;
they cannot be combined with a structured probabilistic model within
the manifold to define a full likelihood relative to the
high-dimensional observations; and they provide only heuristic methods
to evaluate the manifold dimensionality. 
As others have pointed out, they
also make 
 it difficult to extend the manifold definition to
out-of-sample points in a principled way
\cite{Platt05fastmap}.


An established alternative is to construct an explicit probabilistic
model of the functional relationship between low-dimensional manifold
coordinates and each measured dimension of the data, assuming that the
functions instantiate draws from Gaussian-process priors.  The
original \textit{Gaussian process latent variable model} (GP-LVM)
required optimisation of the low-dimensional coordinates, and thus
still did not provide uncertainties on these locations or allow
evaluation of the likelihood of a model over them
\cite{Lawrence03gaussianprocess}; however a recent extension exploits
an auxiliary variable approach to optimise a more general variational
bound, thus retaining approximate probabilistic semantics within the
latent space \cite{Titsias-10}.  The stochastic process model for the
mapping functions also makes it straightforward to estimate the
function at previously unobserved points, thus generalising
out-of-sample with ease. However, the GP-LVM gives up on the intuitive preservation of local
neighbourhood properties that underpin the non-probabilistic methods
reviewed above.  Instead, the expected smoothness or other structure
of the manifold must be defined by the Gaussian process covariance
function, chosen a priori. 

Here, we introduce a new probabilistic model over high-dimensional
observations, low-dimensional embedded locations and locally-linear
mappings between high and low-dimensional linear maps within each
neighbourhood, such that each group of variables is Gaussian
distributed given the other two.  This \textit{locally linear latent
  variable model} (LL-LVM) thus respects the same intuitions as the
common non-probabilistic manifold discovery algorithms, while still
defining a full-fledged probabilistic model.  Indeed, variational
inference in this model follows more directly and with fewer separate
bounding operations than the sparse auxiliary-variable approach used
with the GP-LVM.  Thus, uncertainty in the low-dimensional coordinates
and in the manifold shape (defined by the local maps) is captured
naturally.  A lower bound on the marginal likelihood of the model
makes it possible to select between different latent dimensionalities
and, perhaps most crucially, between different definitions of
neighbourhood, thus addressing an important unsolved issue with
neighbourhood-defined algorithms.
Unlike existing probabilistic frameworks with locally linear models such as
mixtures of factor analysers (MFA)-based and local tangent space analysis
(LTSA)-based methods  \cite{Roweis02globalcoordination, Brand2003, Zhan09},
LL-LVM does not require an additional step to obtain the globally consistent
alignment of low-dimensional local coordinates.\footnote{This is also
  true of one previous MFA-based method \cite{1642659} which finds model parameters
  and global coordinates by variational methods similar to our own.}

This paper is organised as follows. In \secref{lllvm}, we introduce our
generative model, LL-LVM, for which we derive the variational
inference method in \secref{inference}. We briefly describe
out-of-sample extension for LL-LVM and mathematically
describe the dissimilarity between LL-LVM and
GP-LVM at the end of \secref{compare_to_GPLVM}.  
In \secref{experiments}, we demonstrate the
approach  on several real world problems.

{\bf Notation:} 
In the following, a diagonal matrix with entries taken from the vector
$\vv$ is written $\mbox{diag}({\vv})$.
%
%
The vector of $n$ ones is $\vones_n$
and the $n\times n$ identity matrix is $\b{I}_n$.
The Euclidean norm of a vector is $\|\vv\|$, the Frobenius norm of a
matrix is $\|\mathbf{M}\|_F$. 
The Kronecker delta is denoted by $\delta_{ij}$ ($=1$ if $i=j$, and
$0$ otherwise).
The Kronecker product of matrices $\mathbf{M}$ and $\mathbf{N}$ is
$\mathbf{M} \otimes \mathbf{N}$. 
For a random vector $\vw$, we denote the normalisation constant in its
probability density function by $Z_\vw$. 
The expectation of a random vector $\vw$ with respect to a density $q$
is $\langle \vw \rangle_{q}$. 
%


\section{The model: LL-LVM}
\label{sec:lllvm}

Suppose we have $n$ data points $\{\vy_1, \dots, \vy_n\} \subset \mathbb{R}^{d_y}$, and a graph $\graph$ on nodes
$\{1\dots n\}$ with edge set $\edges = \{(i,j) \mid \vy_i \text{ and } \vy_j \text{ are neighbours}\}$.  We assume that
there is a low-dimensional (latent) representation of the high-dimensional data, with coordinates $\{\vx_1, \dots,
\vx_n\} \subset \mathbb{R}^{d_x}$, $d_x < d_y$.  It will be helpful to concatenate the vectors to form $\vy = [
  \vy_1\trp, \dots, \vy_n\trp ]\trp$ and $\vx = [ \vx_1\trp, \dots, \vx_n\trp ]\trp$.

\begin{SCfigure}[50][t] 
\centering
{\includegraphics[width=0.55\textwidth]{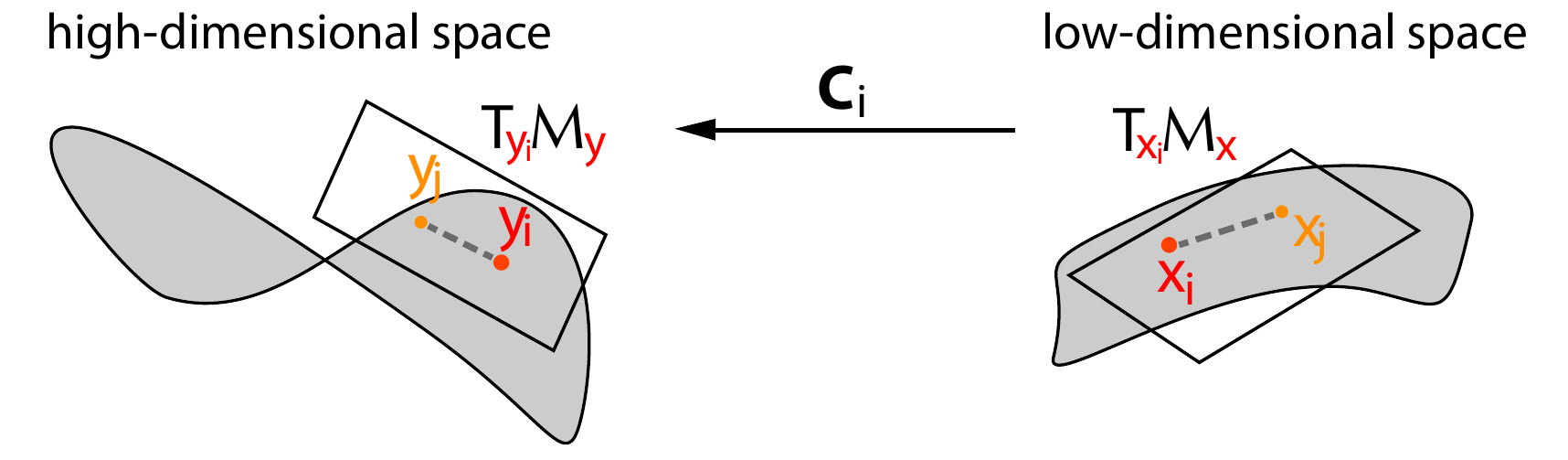}}
\caption{Locally linear mapping $\b{C}_i$ for $i$th data point transforms the tangent space, $T_{\vx_i} \mathcal{M}_{\vx}$ at $\vx_i$ in the low-dimensional space to the tangent space, $T_{\vy_i}\mathcal{M}_{\vy}$ at the corresponding data point $\vy_i$ in the high-dimensional space. A neighbouring data point is denoted by $\vy_j$ and the corresponding latent variable by $\vx_j$.}
\label{fig:Figure1}
\end{SCfigure}

Our key assumption is that the mapping between high-dimensional data and low-dimensional coordinates is  {\it{locally
    linear}} (\figref{Figure1}). The tangent spaces are approximated by $\{ \vy_j - \vy_i\}_{(i,j)\in \edges}$ and $\{\vx_j - \vx_i\}_{(i,j)\in\edges}$, the pairwise differences between the $i$th point and neighbouring points $j$. The matrix ${\mathbf{C}}_i \in \mathbb{R}^{d_y \times d_x}$ at the $i$th point linearly maps those tangent spaces as 
\begin{equation}
\label{eq:modeling_assumption}
\vy_j - \vy_i \approx {\mathbf{C}}_i (\vx_j - \vx_i).
\end{equation} Under this assumption, we aim to find the distribution over the linear maps  $\mathbf{C} =[\mathbf{C}_1, \cdots, \mathbf{C}_n ] \in \mathbb{R}^{d_y \times n d_x}$ and the latent variables $\vx$ that best describe the data likelihood given the graph $\graph$:
\begin{equation}\label{eq:lwr_bnd}
    \log p(\vy|\graph) = \log \int\!\!\!\int p(\vy, \mathbf{C}, \vx|\graph) \, \mathrm{d} \vx\, \mathrm{d}\mathbf{C}.
\end{equation} The joint distribution can be written in terms of priors on  $\mathbf{C}, \vx$ and the likelihood  of  $\vy$ as
\begin{equation}
p(\vy, \mathbf{C}, \vx|\graph) = p(\vy|\mathbf{C}, \vx, \graph) p(\mathbf{C}|\graph) p(\vx|\graph). 
\end{equation}  
In the following, we highlight the essential components the {\it{Locally Linear Latent Variable Model}} (LL-LVM). 
Detailed derivations are given in the Appendix.

\paragraph{Adjacency matrix and Laplacian matrix}  

The edge set of $\graph$ for $n$ data points specifies a $n \times n$ symmetric adjacency matrix $\b{G}$. We write
$\eta_{ij}$ for the $i,j$th element of $\b{G}$, which is 1 if $\vy_j$ and $\vy_i$ are neighbours and 0 if not (including
on the diagonal). The graph Laplacian matrix is then $\b{L} = \mbox{diag}(\b{G}\,\vones_n) - \b{G}$.

\paragraph{Prior on $\vx$}

We assume that the latent variables are zero-centered with a bounded expected scale, and that latent variables
corresponding to neighbouring high-dimensional points are close (in Euclidean distance). Formally, the log prior on the
coordinates is then
\begin{eqnarray}\label{eq:prior_latent}
\log p(\{\vx_1 \dots \vx_n\}|\b{G}, \alpha) = - \tfrac{1}{2} \sum_{i=1}^n (\alpha  \|\vx_i\|^2 +  \sum_{j=1}^n \eta_{ij}\|\vx_i - \vx_j\|^2 ) - \log Z_{\vx}, \nonumber 
\end{eqnarray}  where the parameter $\alpha$ controls the
expected scale ($\alpha > 0$).  This prior can be written as multivariate normal distribution on the concatenated
$\vx$:
\begin{equation}\label{eq:prior_latent_normal}
p(\vx|\b{G}, \alpha) = \Nrm(\mathbf{0}, \b{\Pi}), \; \mbox{ where } \;  \b{\Omega}^{-1} = 2  \b{L}  \otimes  \b{I}_{d_x}, \; \b{\Pi}^{-1} =  \alpha  \b{I}_{n d_x} +  \b{\Omega}^{-1}. \nonumber
\end{equation}

\paragraph{Prior on $\mathbf{C}$}

We assume that the linear maps corresponding to neighbouring points are similar in terms of
Frobenius norm (thus favouring a smooth manifold of low curvature).  This gives
\begin{align}
\log p(\{\mathbf{C}_1 \dots \mathbf{C}_n\} | \b{G}) &= - \frac{\epsilon}{2} \Big\| \sum_{i=1}^n  \mathbf{C}_i
\Big\|_F^2 - \frac{1}{2} \sum_{i=1}^n \sum_{j=1}^n \eta_{ij}\|\mathbf{C}_i -
\mathbf{C}_j\|_F^2 - \log Z_\vc \nonumber  \\
&= - \frac{1}{2} \mbox{Tr} \left[ ( \epsilon \b{J}\b{J}\trp + \b{\Omega}^{-1})
\mathbf{C}\trp  \mathbf{C}  \right] - \log Z_\vc,
\end{align} 
where $\b{J} := \vones_n \otimes \b{I}_{d_x}$.  The second line corresponds to the matrix normal density, giving
$p(\mathbf{C}|\b{G})  = \mathcal{MN}(\b{C}| \b{0},  \b{I}_{d_y},   (\epsilon
\b{J}\b{J}\trp  + \b{\Omega}^{-1})^{-1})$ as the prior on $\mathbf{C}$.
In our implementation, we fix $\epsilon$ to a small value\footnote{$\epsilon$ sets the scale of the average linear map, ensuring the
  prior precision matrix is invertible.}, since the magnitude of the product
$\b{C}_i(\b{x}_i-\b{x}_j)$ is determined by optimising the hyper-parameter $\alpha$ above.
%
%

\begin{SCfigure}[70][t]
\centering
	\centering
	\begin{tikzpicture}[semithick, node distance=4ex, label distance=-0.5ex, ->,
	>=stealth', auto] 
    
	\node[state] (x) {$\vx$}; 

      \node[] [below =5ex of x] (alpha) {$\alpha$};
	
	\node[] [left =6.5ex of x] (G) {$\graph$};
	
	\node[state] [left = 15ex of x] (C) {$\mathbf{C}$};
	
	\node[state, fill=black!25] [below =5ex of G] (y) {$\vy$}; 
    \node[] [right =3ex of y] (V) {$\b{V}$};
	
    \path (V) edge (y);
	\path (alpha) edge (x);
      \path (G) edge (x);
       \path (G) edge (y);
      \path (G) edge (C);
	\path (x) edge(y);
	\path (C) edge(y);
	
	\end{tikzpicture}
\caption{Graphical representation of generative process in
  LL-LVM. Given a dataset, we construct a neighbourhood graph
  $\graph$. The distribution over the latent variable $\vx$ is
  controlled by the graph $\graph$ as well as the parameter
  $\alpha$.  The distribution over the linear map
  $\mathbf{C}$ is also governed by the graph $\graph$. The latent variable
  $\vx$ and the linear map
  $\mathbf{C}$ together determine the data likelihood.}
\label{fig:graphical_model}
\end{SCfigure}
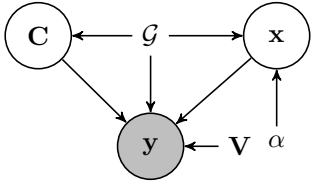

\paragraph{Likelihood}
Under the local-linearity assumption, we penalise the approximation error of \eqref{modeling_assumption}, which yields the log likelihood 
{\small
\begin{equation}
\label{eq:likelihood_Gaussian_in_y}
\log p(\vy|\mathbf{C}, \vx,  \b{V},  \b{G}) = 
- \frac{\epsilon}{2} \| \sum_{i=1}^n \vy_i \|^2
- \tfrac{1}{2} \sum_{i=1}^n
\sum_{j=1}^n\eta_{ij} (\Delta_{\vy_{j,i}} - \mathbf{C}_i
\Delta_{\vx_{j,i}})\trp  \b{V}^{-1} ( \Delta_{\vy_{j,i}} - \mathbf{C}_i
\Delta_{\vx_{j,i}})  - \log Z_\vy , 
\end{equation} 
}%
where $\Delta_{\vy_{j,i}} =\vy_j - \vy_i $ and $\Delta_{\vx_{j,i}} = \vx_j - \vx_i$.\footnote{The $\epsilon$ term
  centers the data and ensures the distribution can be normalised.  It applies in a subspace orthogonal to that modelled
  by $\vx$ and $\b{C}$ and so its value does not affect the resulting manifold model.} 
Thus, $\vy$ is drawn from a multivariate normal distribution given by 
\begin{equation*}
p(\vy|\mathbf{C}, \vx,  \b{V},  \b{G})  = \Nrm(\vmu_\vy, { \b{\Sigma}}_\vy) ,
\end{equation*} 
with $\b{\Sigma}_\vy^{-1} = (\epsilon \vones_n \vones_n\trp) \otimes \b{I}_{d_y} + 2 \b{L} \otimes \b{V}^{-1}$, 
$\vmu_\vy = { \b{\Sigma}}_\vy \ve$,
and $\ve = [\ve_1\trp, \cdots, \ve_n\trp] \trp \in \mathbb{R}^{n d_y}$;  
$\ve_i = - \sum_{j=1}^n \eta_{ji}   \b{V}^{-1} ( \mathbf{C}_j  + \mathbf{C}_i )
\Delta_{\vx_{j,i}}$ .
For computational simplicity, we assume $\b{V}^{-1} = \gamma \b{I}_{d_y}$.
The graphical representation of the generative process underlying the LL-LVM is
given in \figref{graphical_model}.

\section{Variational inference}
\label{sec:inference}

Our goal is to infer the latent variables ($\vx, \mathbf{C}$) as well as the
parameters $\vtheta = \{\alpha,   \gamma \}$ in LL-LVM. We infer them by
maximising the lower bound $\mathcal{L}$ of the marginal likelihood of the
observations
\begin{eqnarray}
\label{eq:variationalLB}
\log p(\vy| \b{G}, \vtheta)
&\geq&  \int\!\!\!\int q(\mathbf{C}, \vx) \; \log \frac{p(\vy, \mathbf{C}, \vx|
\b{G}, \vtheta)}{q(\mathbf{C}, \vx)}\mathrm{d}\vx \mathrm{d} \mathbf{C} \; :=
\mathcal{L}(q(\mathbf{C}, \vx), \vtheta). 
\end{eqnarray} Following the common treatment for computational tractability, we assume the posterior over ($\mathbf{C}, \vx$) factorises as $q(\mathbf{C}, \vx) = q(\vx) q(\mathbf{C})$ \cite{Bishop_06}.
We maximise the lower bound w.r.t.\ $q(\mathbf{C}, \vx)$ and $\vtheta$ by the variational expectation maximization algorithm \cite{Beal_03}, which 
consists of (1) the variational expectation step for computing $q(\mathbf{C}, \vx)$ by
{\small
\begin{align}
\label{eq:Estep_Mbeal_1}
q(\vx) &\propto \exp \left[ \int q(\mathbf{C}) \log p(\vy, \mathbf{C}, \vx|
\b{G}, \vtheta) \mathrm{d} \mathbf{C} \right],  \\
\label{eq:Estep_Mbeal_2}
q(\mathbf{C}) &\propto \exp \left[ \int q(\vx) \log p(\vy, \mathbf{C}, \vx|
\b{G}, \vtheta) \mathrm{d} \mathbf{\vx} \right], 
\end{align}
}%
then (2) the maximization step for estimating $\vtheta$ by $\hat{\vtheta} =
\argmax_{\vtheta} \mathcal{L}(q(\mathbf{C}, \vx), \vtheta)$.
%

\paragraph{Variational-E step}

%
Computing $q(\vx)$ from \eqref{Estep_Mbeal_1} requires rewriting the likelihood in \eqref{likelihood_Gaussian_in_y} as a quadratic function in $\vx$
\begin{align}
p(\vy| \mathbf{C}, \vx, \vtheta,  \b{G}) = \tfrac{1}{\tilde{Z}_\vx}\exp\left[-
\tfrac{1}{2} (\vx \trp \b{A} \vx - 2 \vx\trp \vb)\right], \nonumber
\end{align} 
where the normaliser $\tilde{Z}_\vx$ has all the terms that do not depend on
$\vx$ from \eqref{likelihood_Gaussian_in_y}. 
Let 
$\b{\tilde{L}} := (\epsilon \vones_n \vones_n^\top + 2\gamma \b{L})^{-1} $.
The matrix $\b{A}$ is
given by $\b{A}:=\b{A}_E^{\top} \b{\Sigma}_\b{y} \b{A}_E
= \left[ \b{A}_{ij} \right]_{i,j=1}^n \in \mathbb{R}^{n d_x \times n d_x}$
%
%
where the $i, j$th $d_x \times d_x$ block is 
$\b{A}_{ij} = \sum_{p=1}^n  \sum_{q=1}^n \b{\tilde{L}}(p,q) \b{A}_E(p, i)\trp
\b{A}_E(q, j)$  and  each $i, j$th $(d_y \times d_x)$ block of $\b{A}_E \in
\mathbb{R}^{ n d_y \times n d_x}$  is given by 
$\b{A}_E(i,j) = - \eta_{ij} \b{V}^{-1}  (\b{C}_j + \b{C}_i ) + \delta_{ij}
\left[ \sum_k \eta_{ik} \b{V}^{-1} (\b{C}_k +  \b{C}_i) \right].$
%
The vector $\vb$ is defined as $\vb = [\vb_1\trp, \cdots, \vb_n\trp] \trp \in
\mathbb{R}^{n d_x}$ with the component $d_x$-dimensional vectors given by 
$\vb_i = \sum_{j=1}^n \eta_{ij} ( \b{C}_j\trp \b{V}^{-1} (\vy_i - \vy_j) -
\b{C}_i\trp  \b{V}^{-1} (\vy_j - \vy_i) ).$
The likelihood combined with the prior on $\vx$ gives us the Gaussian posterior
over $\vx$ (i.e., solving \eqref{Estep_Mbeal_1})
%
%
\begin{equation}\label{eq:q_X}
q(\vx) =  \Nrm(\vx|\vmu_\vx,  \b{\Sigma}_\vx), \; \mbox{ where } \;
\b{\Sigma}_\vx^{-1} = \langle  \b{A} \rangle_{q(\mathbf{C})} +
\b{\Pi}^{-1},  \quad \vmu_\vx =  \b{\Sigma}_\vx \langle \vb
\rangle_{q(\mathbf{\bC})}. 
\end{equation}

Similarly, computing $q(\mathbf{C})$ from \eqref{Estep_Mbeal_2} requires
rewriting the likelihood in \eqref{likelihood_Gaussian_in_y} as a quadratic
function in $\mathbf{C}$
%
\begin{equation}
    p(\vy| \mathbf{C}, \vx, \b{G}, \vtheta)= \tfrac{1}{\tilde{Z}_C}
    \exp[- \tfrac{1}{2} \mbox{Tr}( \b{\Gamma} \b{C}\trp \b{C} - 2
\b{C} \trp \b{V}^{-1} \b{H}) ], 
\end{equation} 
where the normaliser $\tilde{Z}_C$ has all the terms that do not depend on
$\mathbf{C}$ from \eqref{likelihood_Gaussian_in_y}, and $\b{\Gamma} := \b{Q}
\tilde{\b{L}} \b{Q}\trp$. The matrix $\b{Q} = [\vq_1
\; \vq_2 \; \cdots \; \vq_n] \in \mathbb{R}^{n d_x \times n}$ where the $j$th
subvector of the $i$th column is $
    \vq_i(j) = \eta_{ij} \b{V}^{-1}(\vx_i - \vx_j)  +  \delta_{ij} \left[
    \sum_k \eta_{ik}  \b{V}^{-1}(\vx_i - \vx_k) \right] \in \mathbb{R}^{d_x}$.
We define $ 
\b{H} = [\b{H}_1, \cdots, \b{H}_n] \in \mathbb{R}^{d_y \times n d_x}$  
whose $i$th block is   $\b{H}_i = \sum_{j=1}^n \eta_{ij} (\vy_j - \vy_i) (\vx_j - \vx_i)\trp
$.


The likelihood combined with the prior on $\b{C}$ gives us the Gaussian posterior
over $\b{C}$  (i.e., solving \eqref{Estep_Mbeal_2})
{\small 
\begin{align}\label{eq:q_C}
    q(\b{C}) = \mathcal{MN} (\vmu_{\b{C}}, \b{I}, \b{\Sigma}_\b{C}),  \text{where } 
    \b{\Sigma}_\b{C}^{-1} := \langle \b{\Gamma} \rangle_{q(\vx)}  +  \epsilon \b{J}
    \b{J}\trp  + \b{\Omega}^{-1}
    \text{ and } 
    \vmu_{\b{C}} =  \b{V}^{-1} \langle \b{H} \rangle_{q(\vx)}  \b{\Sigma}_\b{C}^\top.
\end{align} 
}%
The expected values of $\b{A}, \b{b}, \b{\Gamma}$ and $\b{H}$ are given in the
Appendix.

\begin{figure*}[t]
\centering
\includegraphics[width=0.9\textwidth]{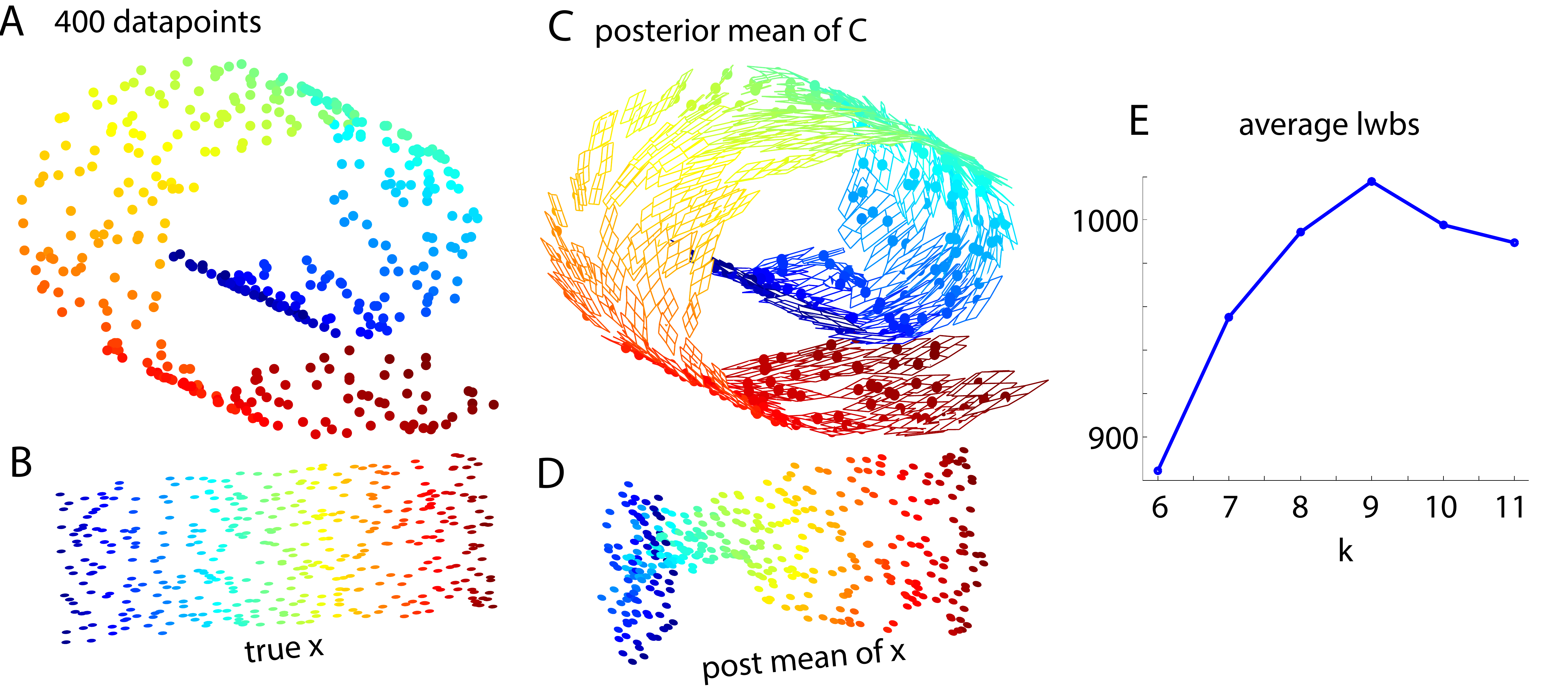}
\caption{A simulated example. \textbf{A}: $400$ data points drawn from Swiss Roll. 
\textbf{B}:  true latent points ($\vx$) in 2D  used for generating the data. 
\textbf{C}:   Posterior mean of $\b{C}$  and \textbf{D}: posterior mean of $\vx$ after 50 EM iterations given $k=9$, which was chosen by maximising the lower bound across different $k$'s.  
\textbf{E}: Average lower bounds as a function of $k$. Each point is an average across $10$ random seeds. }
\label{fig:swissroll}
\end{figure*}

\paragraph{Variational-M step}

We set the parameters by maximising $\mathcal{L}(q(\mathbf{C}, \vx),
\vtheta)$ w.r.t.\ $\vtheta$
which is split into two terms based on dependence on each parameter:
(1) expected log-likelihood for updating $\b{V}$ by $ \arg\max_{\b{V}}
\Em_{q(\vx) q(\mathbf{C})}[\log p(\vy| \mathbf{C}, \vx, \b{V}, \b{G}) ]$; 
and (2) negative KL divergence between the prior and the posterior on $\vx$ for updating $\alpha$ by 
$\arg\max_{\alpha}  \Em_{q(\vx) q(\mathbf{C})} [ \log p(\vx|\b{G}, \alpha) - \log q(\vx) ]$. 
The update rules for each hyperparameter are given in the Appendix.

The full EM algorithm{\footnote{An implementation is available from
\url{http://www.gatsby.ucl.ac.uk/resources/lllvm}.}} starts with an initial value of $\vtheta$. 
In the E-step, given $q(\b{C})$, compute $q(\vx)$ as in \eqref{q_X}. 
Likewise, given $q(\vx)$, compute $q(\b{C})$ as in \eqref{q_C}.
The parameters $\vtheta$ are updated in the M-step 
by maximising \eqref{variationalLB}. The two steps are repeated 
until the variational lower bound in \eqref{variationalLB}  saturates.
To give a sense of how the algorithm works, we visualise fitting results for
a simulated example in \figref{swissroll}. Using the graph constructed
from $3$D observations given different $k$, we run our EM algorithm. The posterior means of $\vx$
and $\b{C}$ given the optimal $k$ chosen by the maximum lower bound resemble the true manifolds in 2D and 3D spaces, respectively.


%

\paragraph{Out-of-sample extension}
\label{sec:out_sample_extension}
In the LL-LVM model one can formulate a computationally efficient out-of-sample extension technique as follows.
Given $n$ data points denoted by $\Dat = \{\vy_1, \cdots, \vy_n \}$,  the variational EM algorithm derived in the previous section converts $\Dat$ into the posterior $q(\vx, \mathbf{C})$: $\Dat \; \mapsto \; q(\vx)q(\mathbf{C})$.
Now, given a new high-dimensional data point $\vy^*$, one can first find the neighbourhood of $\vy^*$ without changing the current neighbourhood graph. Then, it is possible to compute the distributions over the corresponding locally linear map and latent variable $q(\b{C}^*, \vx^*)$ via simply performing the E-step given $q(\vx)q(\mathbf{C})$ (freezing all other quantities the same) as $\Dat \cup \{\vy^*\} \mapsto  q(\vx)q(\mathbf{C})q(\vx^*)q(\b{C}^*)$.

\paragraph{Comparison to GP-LVM}
\label{sec:compare_to_GPLVM}


A closely related probabilistic dimensionality reduction algorithm to LL-LVM is GP-LVM
\cite{Lawrence03gaussianprocess}.  GP-LVM defines the mapping from the latent space to data space using Gaussian
processes. The likelihood of the observations $\b{Y} =[\vy_1, \ldots, \vy_{d_y} ] \in \mathbb{R}^{n \times d_y}$
($\vy_k$ is the vector formed by the $k$th element of all $n$ high dimensional vectors) given latent variables $\b{X}
=[\vx_1, \ldots, \vx_{d_x} ] \in \mathbb{R}^{n \times d_x}$ is defined by $p(\b{Y}|\b{X}) = \prod_{k=1}^{d_y}
\Nrm(\vy_k|\mathbf{0}, \b{K}_{nn} + \beta^{-1} \b{I}_n)$, where the $i,j$th element of the covariance matrix is of the
exponentiated quadratic form: $k(\vx_i, \vx_j) = \sigma^2_f \exp \left[-\tfrac{1}{2}\sum_{q=1}^{d_x} \alpha_q (x_{i, q}
  - x_{j, q} )^2 \right]$ with smoothness-scale parameters $\{ \alpha_q \}$ \cite{Titsias-10}.  In LL-LVM, once we
integrate out $\b{C}$ from \eqref{likelihood_Gaussian_in_y}, we also obtain the Gaussian likelihood given $\vx$,
\begin{align*}
p(\vy|\vx, \b{G}, \vtheta) = \int p(\vy| \mathbf{C}, \vx, \b{G}, \vtheta)
p(\mathbf{C}|\b{G}, \vtheta) \, \mathrm{d}\mathbf{C} = \tfrac{1}{Z_{Y_y}} \exp\left[ -
\tfrac{1}{2} \vy \trp \; \b{K}_{LL}^{-1} \; \vy \right].
\end{align*} 
In contrast to GP-LVM, the precision matrix $\b{K}^{-1}_{LL} = (2\b{L} \otimes \b{V}^{-1}) - ( \b{W} \otimes \b{V}^{-1})
\; {\b{\Lambda}} \; ( \b{W}\trp \otimes \b{V}^{-1})$ depends on the graph Laplacian matrix through $\b{W}$ and
${\b{\Lambda}}$. Therefore, in LL-LVM, the {\it{graph structure}} directly determines the functional form of the
conditional precision.

\section{Experiments}
\label{sec:experiments}

\begin{figure*}[t] 
\centerline{\includegraphics[width=1\textwidth]{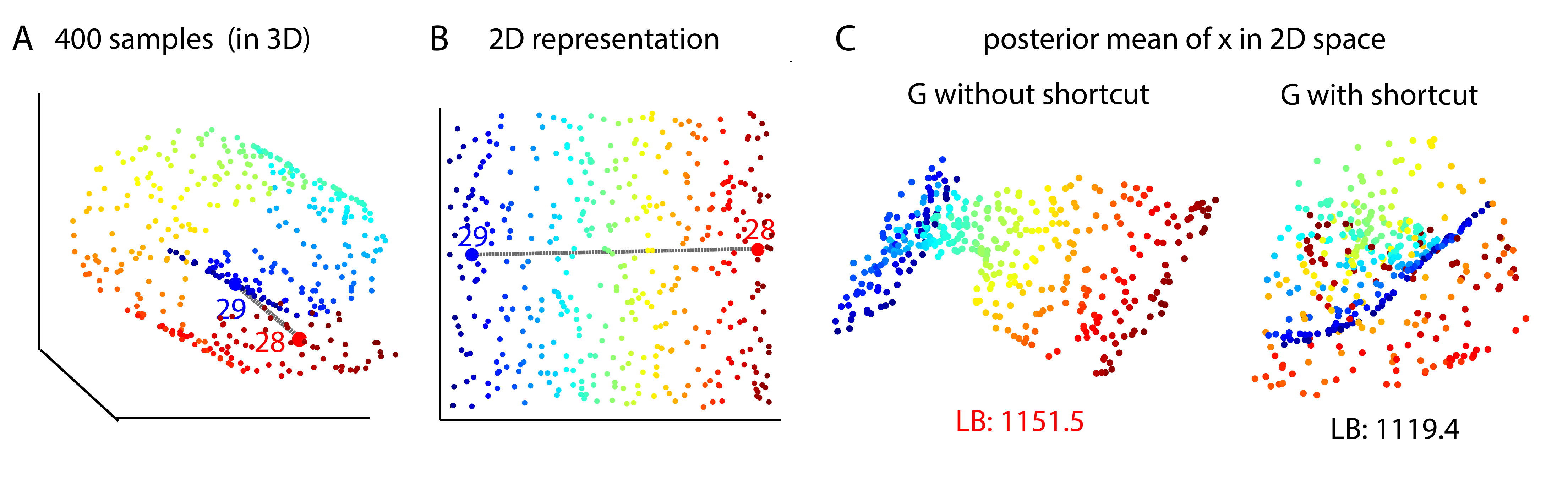}}
\caption{Resolving short-circuiting problems using variational lower
  bound. \textbf{A}: Visualization of $400$ samples drawn from a Swiss
  Roll in 3D space. Points 28 (red) and 29 (blue) are close to each
  other (dotted grey) in 3D. \textbf{B}: Visualization of the $400$
  samples on the latent 2D manifold. The distance between points $28$ and $29$
  is seen to be large. \textbf{C}: Posterior mean of $\vx$ with/without
  short-circuiting the $28$th and the $29$th data points in the graph construction.
  LLLVM achieves a higher lower bound when the shortcut is absent.
  The red and blue parts are mixed in the resulting estimate in 2D space
  (right) when there is a shortcut.
  The lower bound is obtained after 50 EM iterations.
  }
\label{fig:shortcircuiting}
\end{figure*}

\subsection{Mitigating the short-circuit problem}
Like other neighbour-based methods, LL-LVM is sensitive to
misspecified neighbourhoods; the prior, likelihood, and posterior all
depend on the assumed graph.  Unlike other methods, LL-LVM provides a
natural way to evaluate possible short-circuits using the variational
lower bound of \eqref{variationalLB}.  \figref{shortcircuiting} shows
$400$ samples drawn from a Swiss Roll in 3D space
(\figref{shortcircuiting}\textbf{A}). Two points, labelled 28 and 29,
happen to fall close to each other in 3D, but are actually far apart
on the latent (2D) surface (\figref{shortcircuiting}\textbf{B}).  A
k-nearest-neighbour graph might link these, distorting the recovered
coordinates.  However, evaluating the model without this edge (the correct graph)
yields a higher variational bound (\figref{shortcircuiting}\textbf{C}).
Although it is prohibitive to evaluate every possible
graph in this way, the availability of a principled criterion to test
specific hypotheses is of obvious value.

 

%

In the following, we demonstrate LL-LVM on two real datasets: handwritten digits
and climate data.

\subsection{Modelling USPS handwritten digits}
\label{sec:USPS}

\begin{figure}[t] 
\centering
\centerline{\includegraphics[width=1\textwidth]{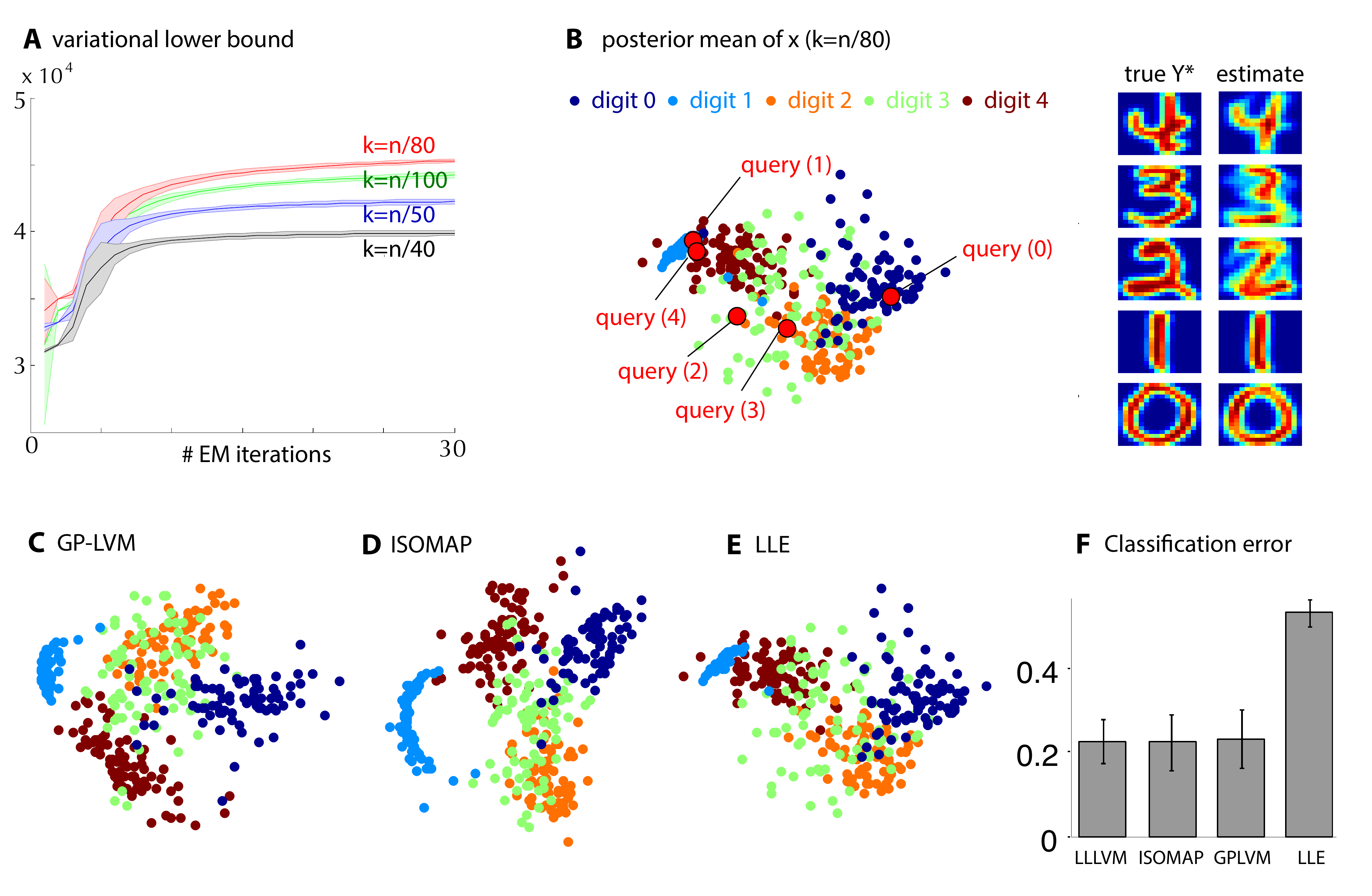}}
\caption{USPS handwritten digit dataset described in \secref{USPS}. \textbf{A}: Mean (in solid) and variance (1 standard
n  deviation shading) of the variational lower bound across 10 different random starts of EM algorithm with different
  $k$'s. The highest lower bound is achieved when $k=n/80$. \textbf{B}: The posterior mean of $\vx$ in 2D. Each digit is
  colour coded. On the right side are reconstructions of $\b{y}^*$ for randomly chosen query points $\b{x}^*$. Using
  neighbouring $\b{y}$ and posterior means of $\b{C}$ we can recover $\b{y}^*$ successfully (see text). \textbf{C}:
  Fitting results by GP-LVM using the same data. \textbf{D}: ISOMAP ($k=30$) and \textbf{E}: LLE ($k$=40). Using the
  extracted features (in 2D), we evaluated a $1$-NN classifier for digit identity with 10-fold cross-validation (the
  same data divided into $10$ training and test sets).  The classification error is shown in \textbf{F}. LL-LVM features
  yield the comparably low error with GP-LVM and ISOMAP.}
\label{fig:USPS}
\end{figure}

As a first real-data example, we test our method on a subset of $80$
samples each of the digits $0, 1, 2, 3, 4$ from the USPS digit
dataset, where each digit is of size $16 \times 16$ (i.e., $n = 400$,
$d_y = 256)$. We follow \cite{Lawrence03gaussianprocess}, and
represent the low-dimensional latent variables in 2D.  

\figref{USPS}\textbf{A} shows variational lower bounds for different
values of $k$, using $9$ different EM initialisations. The posterior
mean of $\vx$ obtained from LL-LVM using the best $k$ is illustrated in
\figref{USPS}\textbf{B}.  \figref{USPS}\textbf{B} also shows
reconstructions of one randomly-selected example of each digit, using
its 2D coordinates $\b{x}^*$ as well as the posterior mean coordinates
$\hat{\b{x}}_i$, tangent spaces $\hat{\b{C}}_i$ and actual images
$\b{y}_i$ of its $k=n/80$ closest neighbours.  The reconstruction is
based on the assumed tangent-space structure of the generative model
(\eqref{likelihood_Gaussian_in_y}), that is:
$\hat{\b{y}}^* = \frac{1}{k} \sum_{i=1}^k \left[ \b{y}_i +
  \hat{\b{C}}_i(\b{x}^* - \hat{\b{x}}_i) \right]$.
A similar process could be used to reconstruct digits at out-of-sample
locations.
Finally, we quantify the relevance of the recovered subspace by
computing the error incurred using a simple classifier to report digit
identity using the 2D features obtained by LL-LVM and various
competing methods (\figref{USPS}\textbf{C-F}).  Classification with
LL-LVM coordinates performs similarly to GP-LVM and ISOMAP ($k=30$), and
outperforms LLE ($k=40$).




\begin{figure}[t]
\centering
 \subfloat[400 weather stations\label{fig:weather_stations}]{
     \includegraphics[width=0.32\textwidth]{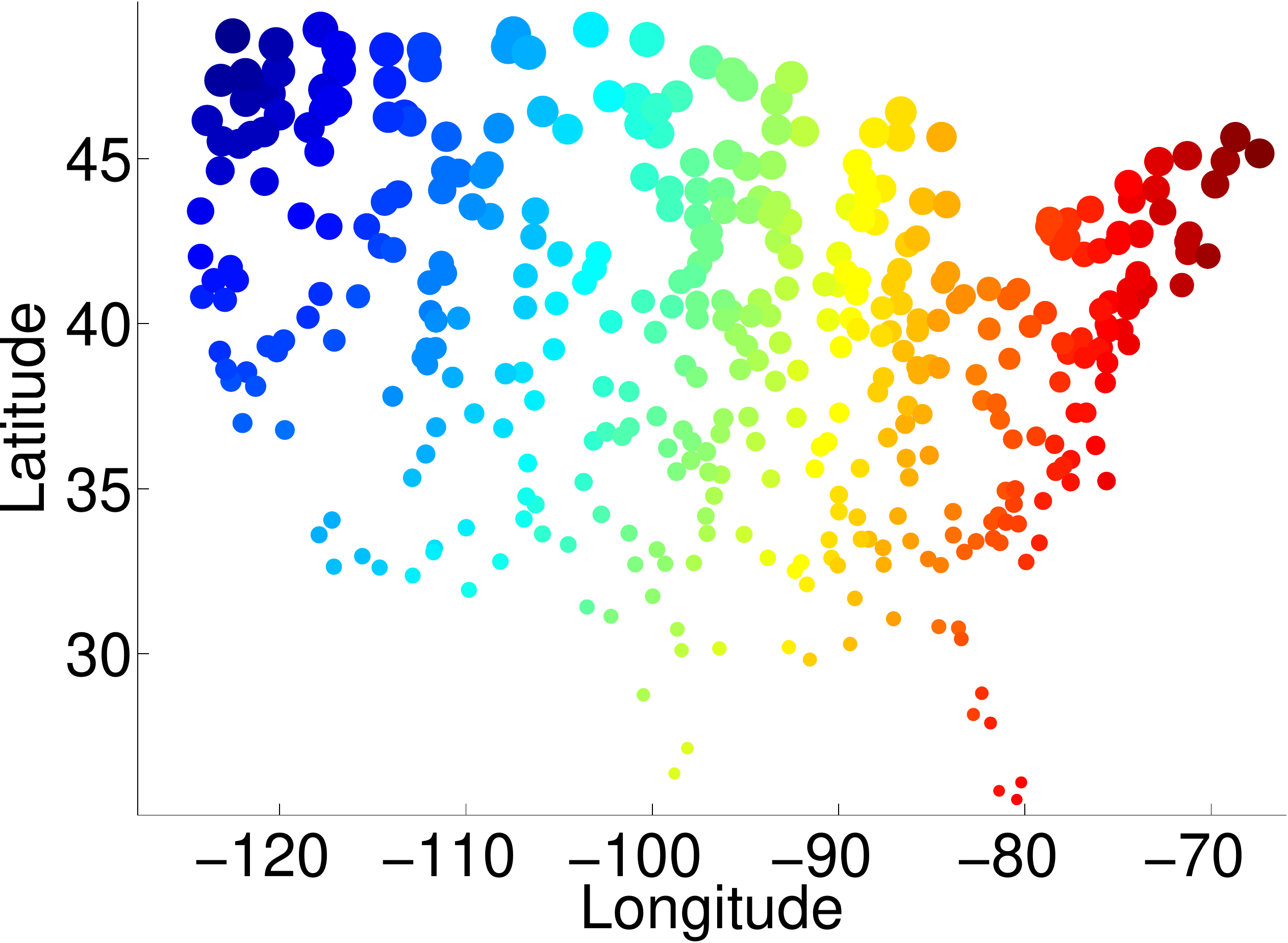}
 }
 \hspace{1mm}
  \subfloat[LLE\label{fig:climate_lle}]{
 \includegraphics[width=0.22\textwidth]{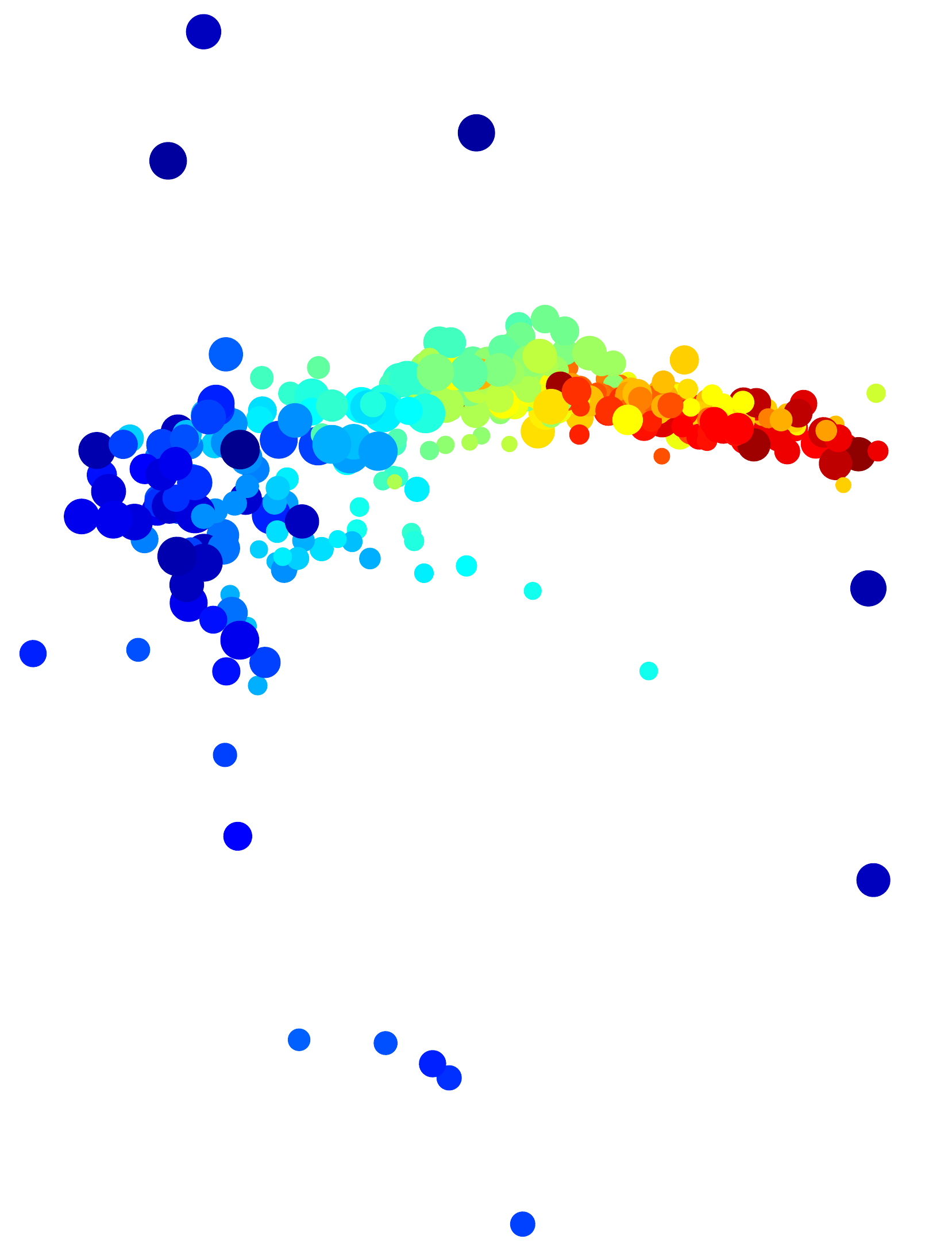}
 } 
\hspace{1mm}
 \subfloat[LTSA\label{fig:climate_ltsa}]{
     \includegraphics[width=0.25\textwidth]{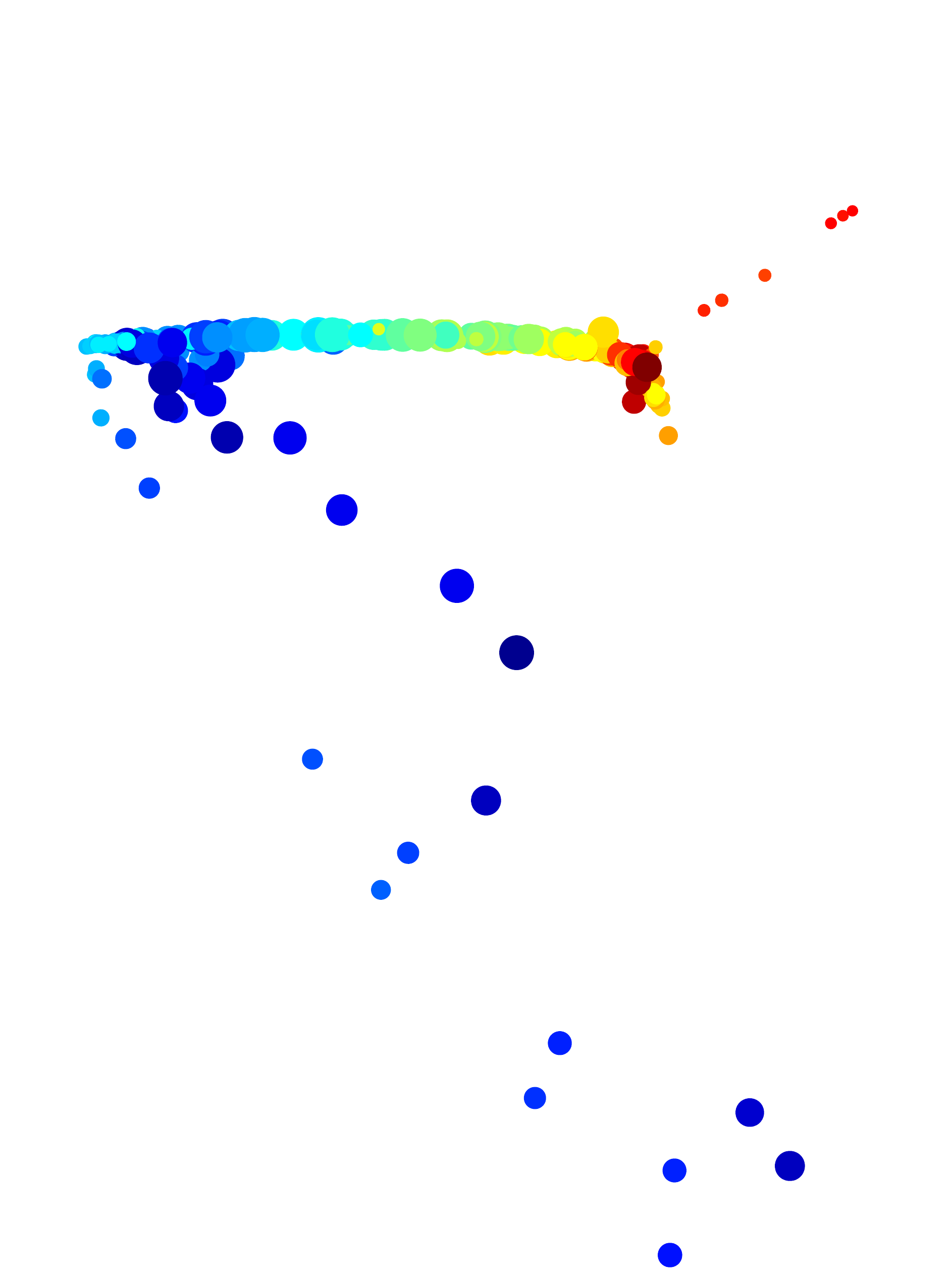}
 }

 \subfloat[ISOMAP\label{fig:climate_isomap}]{
 \includegraphics[width=0.3\textwidth]{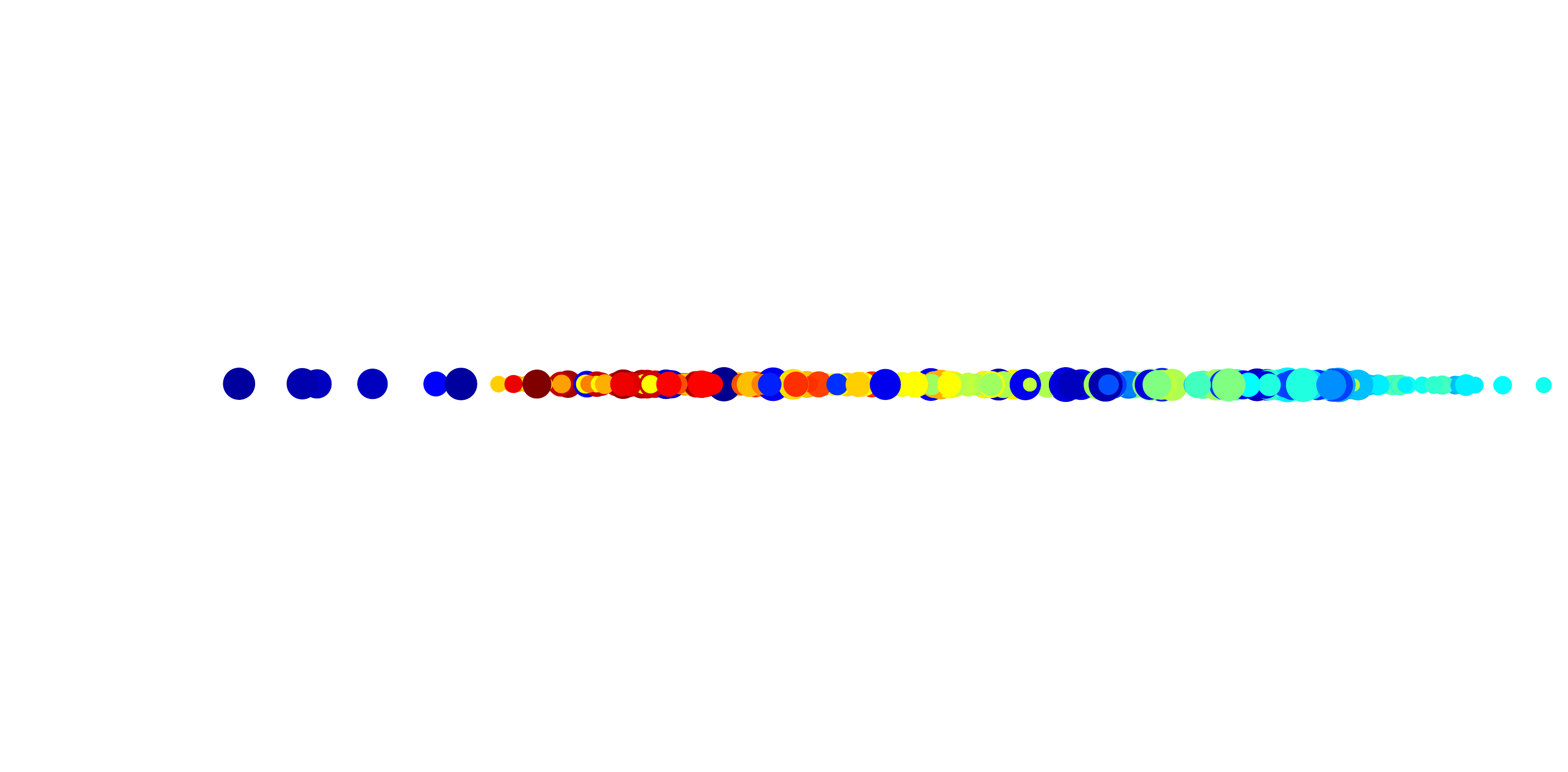}
 }
  \subfloat[GP-LVM\label{fig:climate_gplvm}]{
 \includegraphics[width=0.26\textwidth]{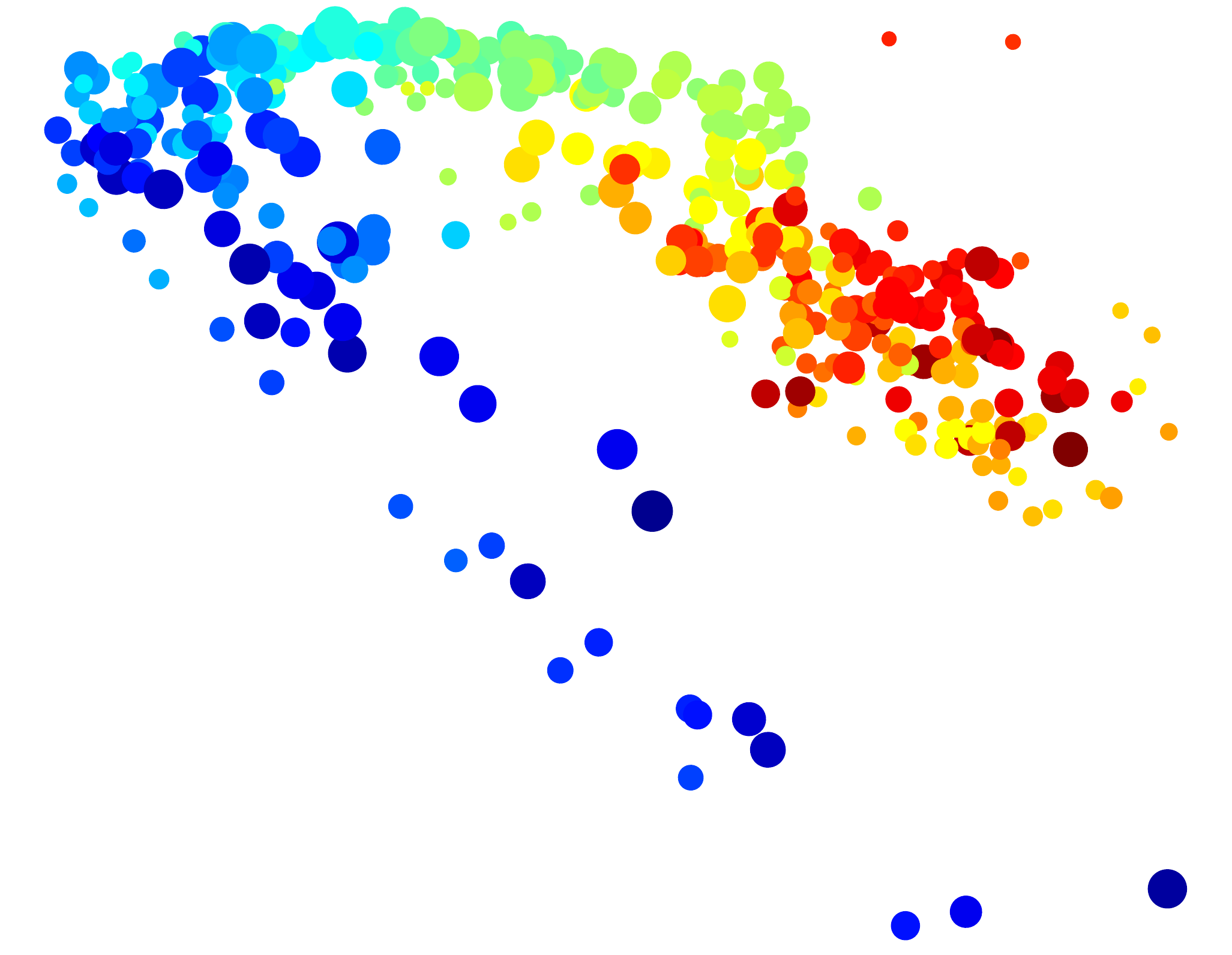}
 }
  \subfloat[LL-LVM\label{fig:climate_lllvm}]{
     \includegraphics[width=0.28\textwidth]{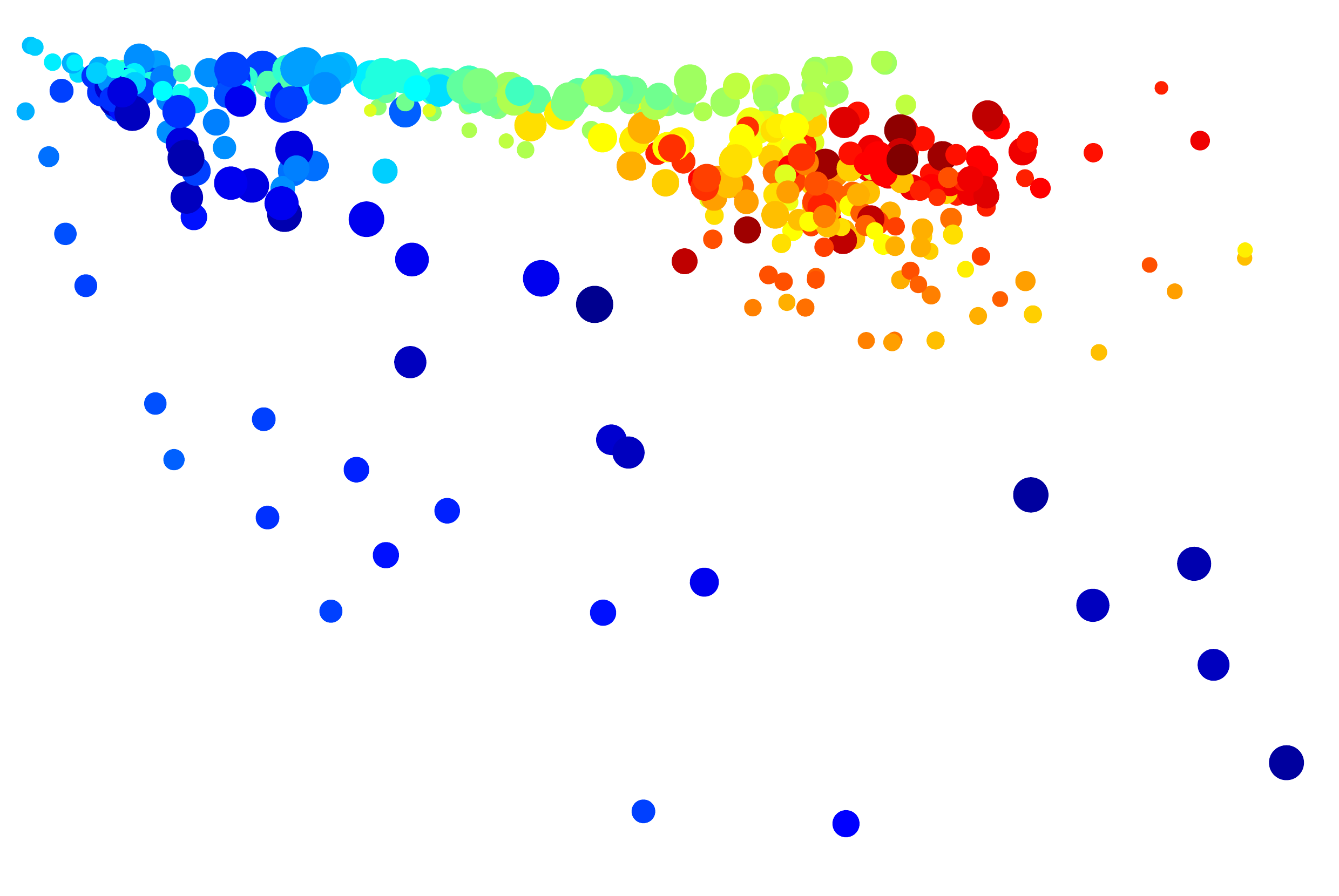}
 }

\caption{Climate modelling problem as described in
  \secref{climate_modelling}. Each example corresponding to a weather
  station is a 12-dimensional vector of monthly precipitation
  measurements. Using only the measurements, the projection obtained
  from the proposed LL-LVM recovers the topological arrangement of the
  stations to a large degree.}
 \label{fig:climate_results}
\end{figure}

\subsection{Mapping climate data}
\label{sec:climate_modelling}
In this experiment, we attempted to recover 2D geographical
relationships between weather stations from recorded monthly
precipitation patterns.  Data were obtained by averaging
month-by-month annual precipitation records from 2005--2014 at 400
weather stations scattered across the US (see
\figref{climate_results}) \footnote{The dataset is made available by
  the National Climatic Data Center at
  \url{http://www.ncdc.noaa.gov/oa/climate/research/ushcn/}. We use
  version 2.5 monthly data \cite{Menne2009}.}.
Thus, the data set comprised 400 12-dimensional vectors.  The goal of
the experiment is to recover the two-dimensional topology of the
weather stations (as given by their latitude and longitude) using only
these 12-dimensional climatic measurements.  As before, we compare the
projected points obtained by LL-LVM with several widely used
dimensionality reduction techniques. 
For the graph-based methods LL-LVM, LTSA, ISOMAP, and LLE, we used
12-NN with Euclidean distance to construct the neighbourhood graph.

The results are presented in \figref{climate_results}.
LL-LVM identified a more geographically-accurate arrangement for the
weather stations than the other algorithms. The fully probabilistic
nature of LL-LVM and GPLVM allowed these algorithms to handle the
noise present in the measurements in a principled way. This contrasts
with ISOMAP which can be topologically unstable
\cite{Balasubramanian2002} i.e.\ vulnerable to short-circuit errors
if the neighbourhood is too large.  Perhaps
coincidentally, LL-LVM also seems to respect local geography more fully
in places than does GP-LVM.



\section{Conclusion}
\label{sec:conclusion}

We have demonstrated a new probabilistic approach to non-linear
manifold discovery that embodies the central notion that local
geometries are mapped linearly between manifold coordinates and
high-dimensional observations.  The approach offers a natural
variational algorithm for learning, quantifies local uncertainty in
the manifold, and permits evaluation of hypothetical neighbourhood
relationships.

In the present study, we have described the LL-LVM model conditioned
on a neighbourhood graph. In principle, it is also possible to extend
LL-LVM so as to construct a distance matrix as in \cite{Lawrence2011},
by maximising the data likelihood.  We leave this as a direction for future
work.

\subsubsection*{Acknowledgments}
The authors were funded by the Gatsby Charitable Foundation.

%
%
%
%

\bibliography{LL_LVM_NIPS_2015}
\bibliographystyle{unsrt}

\newgeometry{lmargin=1in, rmargin=1in}
\clearpage
\newpage
\normalsize
\appendix





\section*{LL-LVM supplementary material}  


\paragraph{Notation}
The vectorized version of a matrix is $\mbox{vec}(\mathbf{M})$.  
We denote an identity matrix of size $m$ with $\b{I}_m$.
Other notations are the same as used in the main text.

\section{Matrix normal distribution}  
\label{sec:matrix_normal}
The matrix normal distribution generalises the standard multivariate normal distribution 
to matrix-valued variables. A matrix  $\b{A} \in \mathbb{R}^{n \times p}$ is said 
to follow a matrix normal distribution $\mathcal{MN}_{n,p}(\mathbf{M}, \mathbf{U}, \mathbf{V})$
with parameters $\b{U}$ and $\b{V}$ if its 
density is given by 
\begin{align}
p(\mathbf{A}\mid\mathbf{M}, \mathbf{U}, \mathbf{V}) = \frac{\exp\left(
    -\frac{1}{2} \, \mathrm{Tr}\left[ \mathbf{V}^{-1} (\mathbf{A} -
    \mathbf{M})^{T} \mathbf{U}^{-1} (\mathbf{A} - \mathbf{M}) \right]
\right)}{(2\pi)^{np/2} |\mathbf{V}|^{n/2} |\mathbf{U}|^{p/2}}.
\end{align}
If $\b{A} \sim \mathcal{MN}(\b{M}, \b{U}, \b{V})$, then $\mathrm{vec}(\b{A})
\sim \mathcal{N}(\mathrm{vec}(\b{M}), \b{V}\otimes \b{U})$, a relationship we will use 
to simplify many expressions.

\section{Matrix normal expressions of priors and likelihood}  

Recall that $\b{G}_{ij} = \eta_{ij}$.
\subsubsection*{Prior on low dimensional latent variables}  
\begin{align}
\log p(\vx|\b{G}, \alpha) &= - \frac{\alpha}{2} \sum_{i=1}^n ||\vx_i||^2 -
\frac{1}{2} \sum_{i=1}^n \sum_{j=1}^n \eta_{ij}||\vx_i - \vx_j||^2 - \log
Z_\vx \\
&= - \frac{1}{2} \log| 2\pi \b{\Pi}| - \frac{1}{2} \vx \trp \b{\Pi}^{-1} \vx,
\end{align}
where 
\begin{align*}
  \b{\Pi}^{-1} &:=  \alpha \b{I}_{n d_x} + \b{\Omega}^{-1},  \\
\b{\Omega}^{-1} &:= 2 \b{L} \otimes \b{I}_{d_x}, \\ 
\b{L} &:= \mathrm{diag}(\b{G}\b{1}) -\b{G}.
\end{align*}
$\b{L}$ is known as a graph Laplacian. It follows that 
$p(\vx|\b{G}, \alpha) = \Nrm(\b{0}, \b{\Pi})$.
The prior covariance $\b{\Pi}$ can be rewritten as
\begin{align}
\b{\Pi}^{-1} &=  \alpha \b{I}_n \otimes \b{I}_{d_x} + 2 \b{L} \otimes \b{I}_{d_x} \\
&= ( \alpha \b{I}_n + 2 \b{L}  ) \otimes \b{I}_{d_x}, \\
\b{\Pi} &=  ( \alpha \b{I}_n + 2 \b{L}  )^{-1} \otimes \b{I}_{d_x}.
\end{align} 
By the relationship of a matrix normal and multivariate normal distributions described 
in \secref{matrix_normal}, the equivalent prior for the matrix $\b{X} = [ \vx_1
\vx_2 \cdots \vx_n ] \in \mathbb{R}^{d_x \times n}$, constructed by  reshaping $\b{x}$, is given by 
\begin{align}
p(\b{X}|\b{G}, \alpha) &= \mathcal{MN}(\b{X}|\b{0},  \; \b{I}_{d_x},  \; ( \alpha \b{I}_n + 2 \b{L}  )^{-1} ).
\end{align}
%

\subsubsection*{Prior on locally linear maps} 
Recall that $\b{C} = [\b{C}_1, \ldots, \b{C}_n ] \in \mathbb{R}^{d_y \times n
d_x}$ where each $\b{C}_i \in \mathbb{R}^{d_y \times d_x}$. We formulate the
log prior on $\b{C}$ as 
\begin{align}
\log p(\mathbf{C}| \b{G}) &= - \frac{\epsilon}{2} || \sum_{i=1}^n  \mathbf{C}_i ||_F^2 - \frac{1}{2} \sum_{i=1}^n \sum_{j=1}^n \eta_{ij}||\mathbf{C}_i - \mathbf{C}_j||_F^2 - \log Z_\vc, \nonumber  \\
&= - \frac{\epsilon}{2} \mbox{Tr} \left( \mathbf{C}\b{J}\b{J}\trp  \mathbf{C} \trp \right) - \frac{1}{2} \mbox{Tr} \left(  \b{\Omega}^{-1} \mathbf{C}\trp  \mathbf{C}  \right) - \log Z_\vc, \mbox{ where } \b{J} := \b{1}_n \otimes \b{I}_{d_x}\nonumber, \\
&= - \frac{1}{2} \mbox{Tr} \left[ ( \epsilon \b{J}\b{J}\trp + \b{\Omega}^{-1}) \mathbf{C}\trp  \mathbf{C}  \right] - \log Z_\vc. 
\end{align} 
In the first line, the first term imposes a constraint that the mean of $\b{C}_i$ should not 
be too large. The second term encourages the the locally linear maps of
neighbouring points $i$ and $j$ to be similar in the sense of the Frobenius
norm. Notice that the last line is in the form of a the log of a matrix normal
density with mean 0 where $Z_\vc$ is given by
\begin{align}
\log Z_\vc = \frac{n d_x d_y}{2} \log |2 \pi | -  \frac{d_y}{2} \log |\epsilon \b{J}\b{J}\trp  + \b{\Omega}^{-1} |
\end{align}
The expression is equivalent to 
\begin{align}
p(\mathbf{C}|\b{G})  &= \mathcal{MN}(\b{C}| \b{0},  \b{I}_{d_y},   ( \epsilon \b{J}\b{J}\trp  + \b{\Omega}^{-1})^{-1}).
\end{align} 
In our implementation, we fix $\epsilon$ to a small value, since the magnitude of 
$\b{C}_i$ and $\b{x}_i$ can be controlled by the hyper-parameter $\alpha$, which is optimized 
in the M-step.

\subsubsection*{Likelihood}

We penalise linear approximation error of the tangent spaces.
Assume that the noise precision matrix is a scaled identify matrix ei.g.,
$\b{V}^{-1} = \gamma \b{I}_{d_y}$.
\begin{align}
\label{eq:likelihood_Gaussian_in_y_appendix}
\log p(\vy|\vx, \b{C}, \b{V}, \b{G}) &= - \frac{\epsilon}{2} || \sum_{i=1}^n  \vy_i ||^2
- \log Z_\vy \\
& -\frac{1}{2} \sum_{i=1}^n \sum_{j=1}^n \eta_{ij} (( \vy_j - \vy_i) - \b{C}_i (
\vx_j - \vx_i))\trp \b{V}^{-1} (( \vy_j - \vy_i) - \b{C}_i ( \vx_j - \vx_i)) , \nonumber \\
&= - \frac{1}{2} (\vy \trp {\Sigma_\vy}^{-1} \vy - 2 \vy \trp \ve + f) - \log Z_\vy, 
\end{align} 
where 
\begin{align}\label{eq:cov_in_y}
\vy &= [\vy_1\trp, \cdots, \vy_n\trp]\trp \in \mathbb{R}^{n d_y} \\
\Sigma_\vy^{-1} &= (\epsilon \b{1}_n \b{1}_n\trp) \otimes \b{I}_{d_y} + 2 \b{L} \otimes \b{V}^{-1}, \\
\ve &= [\ve_1\trp, \cdots, \ve_n\trp] \trp \in R^{n d_y}, \\
\ve_i &= - \sum_{j=1}^n \eta_{ji}  \b{V}^{-1} (\b{C}_j + \b{C}_i) (\vx_j - \vx_i), \\
f &= \sum_{i=1}^n \sum_{j=1}^n \eta_{ij} ( \vx_j - \vx_i)\trp \b{C}_i\trp \b{V}^{-1} \b{C}_i ( \vx_j - \vx_i).
\end{align} 

By completing the quadratic form in $\vy$, we want to write down the likelihood as a multivariate Gaussian
\footnote{The equivalent expression in term of matrix normal distribution for $\b{Y} = [ \vy_1, \vy_2, \cdots,  \vy_n ] \in \mathbb{R}^{d_y \times n}$ 
\begin{align*}
p(\b{Y}|\vx, \b{C}, \gamma, \b{G}) &= \mathcal{MN}(\b{Y} | \b{M}_\vy, \b{I}_{d_y}, (\epsilon \b{1}_n \b{1}_n\trp+ 2 \gamma \b{L} )^{-1}) , \\
\b{M}_\vy &= \b{E} ( \epsilon \b{1}_n \b{1}_n\trp + 2 \gamma \b{L} )^{-1}, 
\end{align*} where $\b{E} = [\ve_1, \cdots, \ve_n ] \in R^{d_y \times n}$. The covariance in \eqref{cov_in_y} decomposes 
\begin{align*}
\Sigma_\vy^{-1} &= (\epsilon \b{1}_n \b{1}_n\trp) \otimes \b{I}_{d_y}  + 2 \b{L} \otimes \b{V}^{-1}, \\
&= (\epsilon \b{1}_n \b{1}_n\trp + 2 \gamma \b{L} )\otimes \b{I}_{d_y}, \\
\Sigma_\vy &= (\epsilon \b{1}_n \b{1}_n\trp + 2 \gamma \b{L} )^{-1} \otimes \b{I}_{d_y}. 
\end{align*}}
:
\begin{align}\label{eq:likelihood_Gaussian_in_y2}
p(\vy|\vx, \b{C}, \b{V}, \b{G}) &=  \Nrm(\vmu_\vy, {\Sigma}_\vy) , \\
\vmu_\vy &= {\Sigma}_\vy \ve.
\end{align} 
By equating \eqref{likelihood_Gaussian_in_y_appendix} with  \eqref{likelihood_Gaussian_in_y2}, we get the normalisation term $ Z_\vy$
\begin{align}
- \frac{1}{2} (\vy \trp {\Sigma_\vy}^{-1} \vy - 2 \vy \trp \ve + f) - \log Z_\vy &= -\frac{1}{2}(\vy-\vmu_\vy)\trp{\Sigma}_\vy^{-1}(\vy-\vmu_\vy) - \frac{1}{2}\log|2\pi {\Sigma}_\vy|, \\
 \log Z_\vy &= \frac{1}{2}(\vmu_\vy\trp{\Sigma}_\vy^{-1} \vmu_\vy -f) + \frac{1}{2}\log|2\pi {\Sigma}_\vy|, \\
 Z_\vy &= \exp(\tfrac{1}{2}(\vmu_\vy\trp{\Sigma}_\vy^{-1} \vmu_\vy -f ))|2\pi {\Sigma}_\vy|^{\frac{1}{2}}, \\
&= \exp(\tfrac{1}{2}(\ve\trp{\Sigma}_\vy \ve -f ))|2\pi {\Sigma}_\vy|^{\frac{1}{2}}.
\end{align} 
Therefore, the normalised log-likelihood  can be written as 
\begin{align}\label{eq:likelihood_normalised}
\log p(\vy|\vx, \b{C}, \b{V}, \b{G}) &= - \frac{1}{2} (\vy \trp {\Sigma_\vy}^{-1} \vy - 2 \vy \trp \ve + \ve\trp{\Sigma}_\vy \ve) - \frac{1}{2}\log |2\pi {\Sigma}_\vy|. 
\end{align} 

\subsubsection*{Convenient form for EM }

For the EM derivation in the next section, it is convenient to write the
exponent term in terms of linear and quadratic functions in $\vx$ and $\b{C}$,
respectively. The linear terms appear in $ \vy \trp \ve$, which we write as 
a linear function in $\vx$ or $\b{C}$
\begin{align} 
\vy \trp \ve &= \vx\trp \vb, \\
&= \mbox{Tr}(\b{C}\trp \b{V}^{-1} \b{H}),
\end{align} 
where 
\begin{align} 
\b{H} &= [\b{H}_1, \cdots, \b{H}_n] \in R^{d_y \times n d_x},  \quad \mbox{ where } \b{H}_i = \sum_{j=1}^n \eta_{ij} (\vy_j - \vy_i) (\vx_j - \vx_i)\trp, \\
\vb &= [\vb_1\trp, \cdots, \vb_n\trp] \trp \in R^{n d_x}, \quad \mbox{ where } \vb_i = \sum_{j=1}^n \eta_{ij} ( \b{C}_j\trp \b{V}^{-1} (\vy_i - \vy_j) -  \b{C}_i\trp  \b{V}^{-1} (\vy_j - \vy_i) ).
\end{align}
The quadratic terms appear in $ \ve\trp{\Sigma}_\vy \ve$, which we write as 
 a quadratic function of $\vx$ or  a quadratic function of $\b{C}$
\begin{align} 
\ve\trp \Sigma_\vy \ve &=  \vx\trp \b{A}_E\trp \Sigma_\vy \b{A}_E \vx  , \\
&=  \mbox{Tr}[ \b{Q} \tilde{\b{L}} \b{Q}\trp \b{C}\trp \b{C}],
\end{align} where the $i, j$th $(d_y \times d_x)$ chunk of $\b{A}_E \in \mathbb{R}^{ n d_y \times n d_x}$  is given by 
\begin{align}
\b{A}_E(i,j) = - \eta_{ij} \b{V}^{-1}  (\b{C}_j + \b{C}_i ) + \delta_{ij} \left[ \sum_k \eta_{ik} \b{V}^{-1} (\b{C}_k +  \b{C}_i) \right].
\end{align} The matrix  $\tilde{\b{L}} = (\epsilon \b{1}_n \b{1}_n\trp + 2 \gamma \b{L})^{-1}$ and 
$\b{Q} = [\vq_1 \; \vq_2 \; \cdots \; \vq_n] \in \mathbb{R}^{n d_x \times n}$ and the $i$th column of this matrix is denoted by $\vq_i \in \mathbb{R}^{nd_x}$.  
The $j$th chunk (of length $d_x$) of the $i$th column is given by 
\begin{align}
\vq_i(j) = \eta_{ij} \b{V}^{-1}(\vx_i - \vx_j)  +  \delta_{ij} \left[ \sum_k \eta_{ik}  \b{V}^{-1}(\vx_i - \vx_k) \right].
\end{align}

\section{Variational inference}  

In LL-LVM, the goal is to infer the latent variables ($\vx, \b{C}$) as well as to learn the
hyper-parameters $\vtheta = \{\alpha,   \gamma \}$. We infer
them by maximising the lower bound of the marginal likelihood of the
observations $\vy$.
\begin{eqnarray}
\log p(\vy| \vtheta, \b{G}) &=& \log  \int  \int p(\vy, \mathbf{C}, \vx| \b{G},
\vtheta) \; \mathrm{d}\vx \; \mathrm{d} \mathbf{C}, \nonumber \\
&\geq& \int  \int \int q(\mathbf{C}, \vx) \; \log \frac{p(\vy, \mathbf{C},
\vx|\b{G}, \vtheta)}{q(\mathbf{C}, \vx)}\mathrm{d}\vx \mathrm{d} \mathbf{C} ,
\nonumber \\
& =& \mathcal{F}(q(\mathbf{C}, \vx), \vtheta). \nonumber
\end{eqnarray}  
For computational tractability, we assume that the posterior over ($\mathbf{C},
\vx$) factorizes as 
\begin{eqnarray}
q(\mathbf{C}, \vx) &=& q(\vx) q(\mathbf{C}).
\end{eqnarray}  where $q(\vx)$ and $q(\b{C})$ are multivariate normal
distributions.

We maximize the lower bound w.r.t.\ $q(\mathbf{C}, \vx)$ and $\vtheta$ by the
variational expectation maximization algorithm, which 
consists of (1) the variational expectation step for determining $q(\mathbf{C}, \vx)$ by
\begin{align}
q(\vx) &\propto \exp \left[ \int q(\mathbf{C}) \log p(\vy, \mathbf{C}, \vx|
\b{G}, \vtheta) \mathrm{d} \mathbf{C} \right],  \\
q(\mathbf{C}) &\propto \exp \left[ \int q(\vx)  \log p(\vy, \mathbf{C}, \vx|
\b{G}, \vtheta) \mathrm{d} \mathbf{\vx} \right],
\end{align}
followed by (2) the maximization step for estimating $\vtheta$, $\hat{\vtheta}
= \argmax_{\vtheta} \mathcal{F}(q(\mathbf{C}, \vx), \vtheta)$.


\subsection{VE step}

\subsubsection{Computing $q(\vx)$}
In variational E-step, we compute $q(\vx)$ by integrating out $\mathbf{C}$ from
the total log joint distribution: 
\begin{eqnarray}
\label{eq:Mstep}
\log q(\vx) &=& \Em_{q(\b{C})} \left[ \log p(\vy, \b{C}, \vx| \b{G},\vtheta) \right] + const, \\
&=&  \Em_{q(\b{C})} \left[ \log p(\vy| \b{C}, \vx, \b{G}, \vtheta) +  \log
p(\vx| \b{G},\vtheta) + \log p(\b{C}| \b{G},\vtheta)  \right] + const.
\end{eqnarray} 
To determine $q(\vx)$, we firstly re-write $p(\vy| \b{C}, \vx, \b{G}, \vtheta)$
as a quadratic function in $\vx$ : 
\begin{align}
\log p(\vy| \b{C}, \vx,  \b{G}, \vtheta)&= 
- \frac{1}{2} ( \vx\trp \b{A}_E\trp \Sigma_\vy \b{A}_E \vx - 2 \vx\trp \vb ) + const, 
\end{align} where 
\begin{align}
\b{A} &:= \b{A}_E\trp \Sigma_\vy \b{A}_E, \\
\b{A} &= \begin{bmatrix}
       \b{A}_{11} & \b{A}_{12} & \cdots & \b{A}_{1n}           \\[0.3em]
        \vdots &           &      \ddots      &   \vdots \\[0.3em]
       \b{A}_{n1} &   \cdots  &  \cdots  & \b{A}_{nn}
\end{bmatrix}\in \mathbb{R}^{n d_x \times n d_x}, \\
\b{A}_{ij} &= \sum_{p=1}^n  \sum_{q=1}^n \b{\tilde{L}}(p,q) \b{A}_E(p, i)\trp \b{A}_E(q, j)
\end{align}
where $\b{\tilde{L}} := (\epsilon \b{1}_n \b{1}_n^\top + 2\gamma \b{L})^{-1} $.
With the likelihood expressed as a quadratic function of $\vx$, 
the log posterior over $\vx$ is given by 
\begin{align}\label{eq:Mstep2}
\log q(\vx) &=  - \frac{1}{2} \Em_{q(\mathbf{C})} \left[ \vx \trp \b{A}  \vx - 2 \vx\trp \vb + \vx \trp \b{\Pi}^{-1} \vx  \right] + const, \\
&=  - \frac{1}{2}  \left[ \vx\trp ( \langle \b{A} \rangle_{q(\mathbf{C})} + \b{\Pi}^{-1}) \vx -2 \vx\trp \langle \vb \rangle_{q(\mathbf{C})} \right] + const, 
\end{align} The posterior over $\vx$ is given by 
\begin{align}\label{eq:Mstep3}
q(\vx) &=  \Nrm(\vx|\vmu_\vx, \b{\Sigma}_\vx),
\end{align} where 
\begin{align}
\b{\Sigma}_\vx^{-1} &= \langle \b{A} \rangle_{q(\mathbf{C})} + \b{\Pi}^{-1}, \\
\vmu_\vx &= \b{\Sigma}_\vx \langle \vb \rangle_{q(\mathbf{C})}.
\end{align} 
Notice that the parameters of $q(\vx)$ depend on the sufficient statistics 
$\langle \b{A} \rangle_{q(\mathbf{C})} $ and $\langle \vb \rangle_{q(\mathbf{C})}$ whose 
explicit forms are given in \secref{suff_A_b}.

\subsubsection{Sufficient statistics $\b{A}$ and  $\b{b}$ for $q(\vx)$}
\label{sec:suff_A_b}

Given the posterior over $\vc$, the sufficient statistics $\langle \b{A}
\rangle_{q(\mathbf{C})} $ and $\langle \vb \rangle_{q(\mathbf{C})}$  necessary
to characterise $q(\vx)$ are computed as following: 
\begin{align}
\langle \b{A}_{ij} \rangle_{q(\vc)} &= \sum_{p=1}^n  \sum_{q=1}^n \b{\tilde{L}}(p,q) \langle \b{A}_E(p, i)\trp \b{A}_E(q, j) \rangle_{q(\vc)}, \\
&= \gamma^2  \sum_{p=1}^n  \sum_{q=1}^n  \b{\tilde{L}}(p,q)  \langle ( -\eta_{pi} (\b{C}_p + \b{C}_i) + \delta_{pi} \sum_k \eta_{pk} (\b{C}_k + \b{C}_p))\trp( -\eta_{qj} (\b{C}_q + \b{C}_j) + \delta_{qj} \sum_{k'} \eta_{qk'} (\b{C}_{k'} + \b{C}_q))  \rangle_{q(\vc)}  \nonumber \\
&=  \gamma^2  \sum_{p=1}^n  \sum_{q=1}^n  \b{\tilde{L}}(p,q) ( \; \eta_{pi} \eta_{qj} \langle \b{C}_p\trp \b{C}_q + \b{C}_p\trp \b{C}_j + \b{C}_i\trp \b{C}_q + \b{C}_i\trp \b{C}_j \rangle_{q(\vc)}   \nonumber \\
&\qquad \qquad \qquad \qquad - \eta_{pi} \delta_{qj} \sum_{k'} \eta_{qk'}  \langle \b{C}_p\trp \b{C}_{k'} + \b{C}_p\trp \b{C}_{q} + \b{C}_i\trp \b{C}_{k'} + \b{C}_i\trp \b{C}_{q} \rangle_{q(\vc)} \nonumber \\
&\qquad \qquad \qquad \qquad - \eta_{qj} \delta_{pi} \sum_{k} \eta_{pk}  \langle \b{C}_k\trp \b{C}_{q} + \b{C}_k\trp \b{C}_{j} + \b{C}_p\trp \b{C}_{q} + \b{C}_p\trp \b{C}_{j} \rangle_{q(\vc)} \nonumber \\
&\qquad \qquad \qquad \qquad + \delta_{pi} \delta_{qj} \sum_{k}\sum_{k'} \eta_{pk}\eta_{qk'}  \langle \b{C}_k\trp \b{C}_{k'} + \b{C}_k\trp \b{C}_{q} + \b{C}_p\trp \b{C}_{k'} + \b{C}_p\trp \b{C}_{q} \rangle_{q(\vc)} \; ) \nonumber
\end{align} 
Thanks to the delta function, the last three terms above are non-zero only when $p=i$ and $q=j$. Therefore, we can replace $p$ with $i$, and $q$ with $j$, which simplifies the above as
\begin{align}
&\gamma^2  \sum_{p=1}^n  \sum_{q=1}^n  \b{\tilde{L}}(p,q)  \; \eta_{pi} \eta_{qj} \langle \b{C}_p\trp \b{C}_q + \b{C}_p\trp \b{C}_j + \b{C}_i\trp \b{C}_q + \b{C}_i\trp \b{C}_j \rangle_{q(\vc)}   \nonumber \\
&- \gamma^2  \sum_{p=1}^n \sum_{k'}^n  \b{\tilde{L}}(p,j)   \eta_{pi} \eta_{jk'}  \langle \b{C}_p\trp \b{C}_{k'} + \b{C}_p\trp \b{C}_{j} + \b{C}_i\trp \b{C}_{k'} + \b{C}_i\trp \b{C}_{j} \rangle_{q(\vc)} \nonumber \\
& - \gamma^2  \sum_{q=1}^n \sum_{k} \b{\tilde{L}}(i,q)  \eta_{qj}   \eta_{ik}  \langle \b{C}_k\trp \b{C}_{q} + \b{C}_k\trp \b{C}_{j} + \b{C}_i\trp \b{C}_{q} + \b{C}_i\trp \b{C}_{j} \rangle_{q(\vc)} \nonumber \\
& + \gamma^2 \b{\tilde{L}}(i,j)   \sum_{k}\sum_{k'} \eta_{ik}\eta_{jk'}  \langle \b{C}_k\trp \b{C}_{k'} + \b{C}_k\trp \b{C}_{j} + \b{C}_i\trp \b{C}_{k'} + \b{C}_i\trp \b{C}_{j} \rangle_{q(\vc)} \; \nonumber
\end{align} 
We can make the equation above even simpler by replacing $k'$ with $q$ (second line), $k$ with $p$ (third line), and both $k$ and $k'$ with $p$ and $q$ (fourth line), which gives us
\begin{align}
\langle \b{A}_{ij} \rangle_{q(\vc)}&=\gamma^2  \sum_{p=1}^n  \sum_{q=1}^n  [ \b{\tilde{L}}(p,q) -  \b{\tilde{L}}(p,j) - \b{\tilde{L}}(i,q) + \b{\tilde{L}}(i,j) ]   \; \eta_{pi} \eta_{qj} \langle \b{C}_p\trp \b{C}_q + \b{C}_p\trp \b{C}_j + \b{C}_i\trp \b{C}_q + \b{C}_i\trp \b{C}_j \rangle_{q(\vc)}.
\end{align}

For $\vb_i$, we have
\begin{align}
\langle \vb_i  \rangle_{q(\vc)}&= \gamma \sum_{j=1}^n \eta_{ij} ( \langle \b{C}_j \rangle_{q(\vc)}\trp (\vy_i - \vy_j) -  \langle \b{C}_i\rangle_{q(\vc)} \trp  (\vy_j - \vy_i) ), \\
\mbox{where
} \quad \quad & \\
\langle \b{C}_i \rangle_{q(\vc)} 
&= \mbox{i-th chunk of } \vmu_\b{C}, \; \mbox{where each chunk is } (d_y \times d_x) \\
\langle \b{C}_i\trp  \b{C}_j \rangle_{q(\vc)} 
&= \mbox{(i,j)-th} (d_x \times d_x) \mbox{ chunk of }  dy \b{\Sigma}_\b{C} + \langle \b{C}_i \rangle_{q(\vc)}\trp \langle \b{C}_j \rangle_{q(\vc)}, 
\end{align}
\subsubsection{Computing $q(\b{C})$}
Next, we compute $q(\mathbf{C})$ by integrating out $\vx$ from the total log joint distribution:
\begin{align}
\label{eq:Mstep_C}
\log q(\mathbf{C}) &= \Em_{q(\vx)} \left[ \log p(\vy, \b{C}, \b{X}| \b{G}, \vtheta) \right] + const, \\
&=  \Em_{q(\vx)} \left[ \log p(\vy| \b{C}, \vx, \b{G}, \vtheta) +
\log p(\vx| \b{G},\vtheta) \right]+ \log p(\mathbf{C}| \b{G},\vtheta) + const.
\end{align} 
We re-write $p(\vy| \mathbf{C}, \vx, \b{G}, \vtheta)$ as a quadratic function in $\mathbf{C}$:
\begin{align}
\log p(\vy| \mathbf{C}, \vx, \b{G}, \vtheta)&= 
- \frac{1}{2} \mbox{Tr}( \b{Q} \tilde{\b{L}} \b{Q}\trp \b{C}\trp \b{C} - 2 \b{C} \trp \b{V}^{-1} \b{H}) + const, 
\end{align} where
\begin{align*}
\b{\Gamma} &:= \b{Q} \tilde{\b{L}} \b{Q}\trp, \\
\b{\Gamma} &= \begin{bmatrix}
       \b{\Gamma}_{11} & \b{\Gamma}_{12} & \cdots & \b{\Gamma}_{1n}           \\[0.3em]
        \vdots &           &      \ddots      &   \vdots  \\[0.3em]
       \b{\Gamma}_{n1} &   \cdots  &  \cdots  & \b{\Gamma}_{nn}
\end{bmatrix}
\end{align*}
\begin{align}
\b{\Gamma}_{ij} &=\sum_{k=1}^n  \sum_{k'=1}^n \b{\tilde{L}}(k,k') \vq_k(i) \vq_{k'}(j)\trp.
\end{align}
The log posterior over $\mathbf{C}$ is given by 
\begin{align}
\log q(\mathbf{C}) 
&=  - \frac{1}{2} \mbox{Tr}\left[ \langle \b{\Gamma} \rangle_{q(\vx)}  \mathbf{C}\trp \mathbf{C} - 2 \mathbf{C}\trp \b{V}^{-1} \langle \b{H} \rangle_{q(\vx)} +  (\epsilon \b{J} \b{J}\trp  + \b{\Omega}^{-1}) \mathbf{C}\trp  \mathbf{C}\right] + const, \nonumber
\end{align} 
The posterior over $\b{C}$ is given by 
\begin{align}
\b{\Sigma}^{-1}_{\vc} &=(\langle \b{\Gamma}  \rangle_{q(\vx)}  +  \epsilon \b{J} \b{J}\trp  + \b{\Omega}^{-1} ) \otimes \b{I}, \\
&= \b{\Sigma}_\b{C}^{-1} \otimes \b{I}, \quad \mbox{where } \b{\Sigma}_\b{C}^{-1} := \langle \b{\Gamma} \rangle_{q(\vx)}  +  \epsilon \b{J} \b{J}\trp  + \b{\Omega}^{-1}  \\
\vmu_{\b{C}} &=  \b{V}^{-1} \langle \b{H} \rangle_{q(\vx)}  \b{\Sigma}_\b{C}\trp.
\end{align} 
Therefore, the approximate posterior over $\b{C}$ is given by 
\begin{align}
q(\b{C}) &= \mathcal{MN} (\vmu_{\b{C}}, \b{I}, \b{\Sigma}_\b{C}). 
\end{align}   
The parameters of $q(\b{C})$ depend on the sufficient statistics 
$\langle \b{\Gamma} \rangle_{q(\vx)} $ and $\langle \b{H} \rangle_{q(\vx)}$ which 
are given in \secref{suff_gam_H}.

\subsubsection{Sufficient statistics $\b{\Gamma}$ and  $\b{H}$}
\label{sec:suff_gam_H}

Given the posterior over $\vx$, the sufficient statistics 
$\langle \b{\Gamma} \rangle_{q(\vx)} $ and $\langle \b{H} \rangle_{q(\vx)}$ 
necessary to characterise $q(\b{C})$ are computed as follows.
Similar to $\langle \b{A} \rangle$, we can simplify $\langle \b{\Gamma}_{ij} \rangle_{q(\vx)}$ as
\begin{align}
\langle \b{\Gamma}_{ij} \rangle_{q(\vx)} &=\gamma^2 \sum_{k=1}^n  \sum_{k'=1}^n
[ \b{\tilde{L}}(k,k') -  \b{\tilde{L}}(k,j) - \b{\tilde{L}}(i,k') +
\b{\tilde{L}}(i,j) ]   \; \eta_{ki} \eta_{k'j}\langle \vx_k \vx_{k'}\trp -
\vx_k \vx_j\trp - \vx_i \vx_{k'}\trp + \vx_i \vx_j\trp \rangle_{q(\vx)}.
\label{eq:egam_ij}
\end{align}

For $\langle \b{H}_i \rangle_{q(\vx)} $, we have
\begin{align}
\langle \b{H}_i \rangle_{q(\vx)} &= \sum_{j=1}^n \eta_{ij} \langle (\vy_j - \vy_i) (\vx_j - \vx_i)\trp \rangle_{q(\vx)}, \\
&=  \sum_{j=1}^n \eta_{ij} (\vy_j \langle \vx_j \rangle_{q(\vx)}\trp - \vy_j \langle \vx_i \rangle_{q(\vx)}\trp - \vy_i \langle \vx_j \rangle_{q(\vx)}\trp +  \vy_i \langle \vx_i \rangle_{q(\vx)}\trp), 
\end{align} where $\langle \vx_i  \vx_j \trp  \rangle_{q(\vx)} = \b{\Sigma}_{\vx}^{(ij)} + \langle \vx_i \rangle_{q(\vx)}\langle \vx_j \rangle_{q(\vx)}\trp$ and $ \b{\Sigma}_{\vx}^{(ij)} = \mbox {cov}(\vx_i, \vx_j)$.

\subsection{VM step}
We set the parameters $\vtheta = (\alpha, \gamma)$ by maximising the
free energy w.r.t. $\vtheta$:
\begin{align}
\label{eq:free_energy}
\hat{\vtheta} &= \arg\max_\vtheta \Em_{q(\vx) q(\b{C}) }[\log p(\vy, \b{C}, \vx| \b{G}, \vtheta)  - \log q(\vx, \b{C})] , \nonumber \\
&= \arg\max_\vtheta \Em_{q(\vx) q(\b{C}) }[\log p(\vy| \b{C}, \vx, \b{G}, \vtheta)  + \log p(\b{C}|\b{G}, \vtheta) + \log p(\vx|\b{G}, \vtheta) - \log q(\vx) - \log q(\b{C})].
\end{align} 
Once we update all the parameters, we achieve the following lower bound:
\begin{equation}\label{eq:lwb}
\mathcal{L}(q(\vx, \b{C}), \hat{\vtheta}) = \Em_{q(\vx) q(\b{C}) }[\log p(\vy| \b{C}, \vx, \b{G}, \hat{\vtheta})] -D_{KL}(q(\b{C})||p(\b{C}|\b{G})) -D_{KL}(q(\vx)||p(\vx|\b{G}, \hat{\vtheta})). 
\end{equation}

\subsubsection*{Update for $\gamma$}
Recall that the precision matrix in the likelihood term is $\b{V}^{-1} = \gamma
\b{I}_{d_y}$. For updating $\gamma$, it is sufficient to consider the log conditional
likelihood integrating out $\vx, \mathbf{C}$:
\begin{align}
 \Em_{q(\vx) q(\mathbf{C})}[\log p(\vy| \mathbf{C}, \vx, \b{G}, \vtheta) ] &=   \Em_{q(\vx) q(\mathbf{C})} \left[ - \frac{1}{2} \mbox{Tr}(\b{\Gamma} \mathbf{C}\trp \mathbf{C} - 2 \mathbf{C} \trp \b{V}^{-1} \b{H})  -  \frac{1}{2} \vy\trp \Sigma_\vy^{-1} \vy -\frac{1}{2}\log |2\pi \Sigma_\vy| \right], 
\end{align} 
which is
\begin{align*}
& - \frac{1}{2} \Em_{q(\mathbf{C})} \mbox{Tr}( \langle \b{\Gamma} \rangle_{q(\vx)} \mathbf{C}\trp  \mathbf{C} - 2 \mathbf{C} \trp \b{V}^{-1} \langle \b{H} \rangle_{q(\vx)}) -  \frac{1}{2} \vy\trp \Sigma_\vy^{-1} \vy  -\frac{1}{2}\log |2\pi \Sigma_\vy|, \nonumber \\
&=   - \frac{1}{2} \Em_{q(\mathbf{C})}  [ \vc\trp (\langle \b{\Gamma} \rangle_{q(\vx)} \otimes \b{I}_{d_y}) \vc - 2 \vc\trp \mbox{vec}(\b{V}^{-1} \langle \b{H} \rangle_{q(\vx)}) ] -  \frac{1}{2} \vy\trp \Sigma_\vy^{-1} \vy  -\frac{1}{2}\log |2\pi \Sigma_\vy|, \nonumber \\
&= - \frac{d_y}{2} \mbox{Tr} (\langle \b{\Gamma} \rangle_{q(\vx) } \b{\Sigma}_\b{C})  - \frac{1}{2} \mbox{Tr}(\langle \b{\Gamma} \rangle_{q(\vx)}\vmu_{\b{C}}\trp \vmu_{\b{C}})+  \gamma \mbox{Tr} (\vmu_{\b{C}}\trp \langle \b{H} \rangle_{q(\vx)} ) -  \frac{1}{2} \vy\trp \Sigma_\vy^{-1} \vy  -\frac{1}{2}\log |2\pi \Sigma_\vy|.
\end{align*} The log determinant term is further simplified as 
\begin{align}
 -\frac{1}{2}\log |2\pi \Sigma_\vy| &= - \frac{n d_y}{2} \log (2\pi) + \frac{d_y}{2} \log | \epsilon \b{1}_n \b{1}_n\trp + 2 \gamma \b{L}|.
\end{align} 

We denote the objective function for updating $\gamma$ by $l(\gamma)$, which consists of all the terms that depend on $\gamma$ above 
\begin{align*}
l(\gamma)&= - \frac{1}{2} \mbox{Tr} (\langle \b{\Gamma} \rangle_{q(\vx) } (d_y
\b{\Sigma}_\b{C} + \vmu_{\b{C}}\trp \vmu_{\b{C}}))+  \gamma \mbox{Tr}
(\vmu_{\b{C}}\trp \langle \b{H} \rangle_{q(\vx)} ) \\ 
& -\frac{1}{2} \vy\trp (
(\epsilon \b{1}_n \b{1}_n\trp + 2 \gamma \b{L}) \otimes \b{I}_{dy}) \vy 
+ \frac{d_y}{2} \log | \epsilon \b{1}_n \b{1}_n\trp + 2 \gamma \b{L}|, \\
&= l_1(\gamma)+ l_2(\gamma) + l_3(\gamma) + l_4(\gamma), 
\end{align*} where each term is given below. 
From the definition of $\b{\Gamma}  = \b{Q} \tilde{\b{L}} \b{Q}\trp$, we rewrite the first term above as
\begin{align*}
l_1(\gamma) &=  - \frac{1}{2}  \mbox{Tr} (\langle \b{Q} \tilde{\b{L}} \b{Q}\trp \rangle_{q(\vx) } (d_y \b{\Sigma}_\b{C} + \vmu_{\b{C}}\trp \vmu_{\b{C}})).
 \end{align*} We separate $\gamma$ from $ \b{Q}$ and plug in the definition of $ \tilde{\b{L}}$, which gives us
 \begin{align*}
 l_1(\gamma) &= - \frac{1}{2} \gamma^2 \mbox{Tr} (\langle \hat{\b{Q}} \tilde{\b{L}} \hat{\b{Q}}\trp \rangle_{q(\vx) } (d_y \b{\Sigma}_\b{C} + \vmu_{\b{C}}\trp \vmu_{\b{C}})), 
 \end{align*} where the $j$th chunk (of length $d_x$) of $i$th column of $\hat{\b{Q}} \in \mathbb {R}^{nd_x \times n}$ is given by 
$\hat{\vq}_i(j) = \eta_{ij} (\vx_i - \vx_j) + \delta_{ij} [ \sum_k \eta_{ik} (\vx_i - \vx_k) ].$
We can explicitly write down $\tilde{\b{L}}$ in terms of  $\gamma$ using orthogonality of singular vectors between $\epsilon \b{1}_n \b{1}_n\trp$ and $2 \gamma \b{L}$, where we denote the singular decomposition of $\b{L} = {\b{U}_L} {\b{D}_L} {\b{V}_L}\trp$
  \begin{align*} 
  \tilde{\b{L}} &= (\epsilon \b{1}_n \b{1}_n\trp + 2 \gamma \b{L})^{-1}, \\
  &:={\b{V}_L} 
  \begin{bmatrix}
       0 & 0 & \cdots & 0           \\[0.3em]
       0 &           &      \ddots      &   \vdots  \\[0.3em]
       0 &   \cdots  &  \cdots  &  \frac{1}{\epsilon n}
\end{bmatrix}{\b{U}_L} \trp +
\frac{1}{2 \gamma}
{\b{V}_L} 
  \begin{bmatrix}
       \frac{1}{\b{D}_L(1,1)} & 0 & \cdots & 0           \\[0.3em]
       0 &    \frac{1}{\b{D}_L(2,2)}       &      \ddots      &   \vdots  \\[0.3em]
       0 &   \cdots  &  \cdots  & 0
\end{bmatrix}{\b{U}_L} \trp, \\
&=   \tilde{\b{L}}_\epsilon + \frac{1}{2\gamma}   \tilde{\b{L}}_L 
 \end{align*} 
 Hence, $ l_1(\gamma) $ is given by  
 \begin{align*}
 l_1(\gamma) &= - \frac{1}{2} \gamma^2 \mbox{Tr} ( \langle  \hat{\b{Q}}
 \tilde{\b{L}}_\epsilon \hat{\b{Q}}\trp \rangle_{q(\vx) } (d_y \b{\Sigma}_\b{C}
 + \vmu_{\b{C}}\trp \vmu_{\b{C}})) - \frac{\gamma}{4} \mbox{Tr} ( \langle
 \hat{\b{Q}}  \tilde{\b{L}}_L \hat{\b{Q}}\trp \rangle_{q(\vx) } (d_y
 \b{\Sigma}_\b{C} + \vmu_{\b{C}}\trp \vmu_{\b{C}})). 
 \end{align*} 
 Let $\langle \b{\Gamma}_\epsilon \rangle := \langle  \hat{\b{Q}}
 \tilde{\b{L}}_\epsilon \hat{\b{Q}}\trp \rangle_{q(\vx) }$.
 Similar to Eq.~\ref{eq:egam_ij}, we have
\begin{align*}
    \langle \b{\Gamma}_{\epsilon, ij} \rangle &=\gamma^2 \sum_{k=1}^n  \sum_{k'=1}^n
[ \b{\tilde{L}_\epsilon}(k,k') -  \b{\tilde{L}_\epsilon}(k,j) - \b{\tilde{L}_\epsilon}(i,k') +
\b{\tilde{L}_\epsilon}(i,j) ]   \; \eta_{ki} \eta_{k'j}\langle \vx_k \vx_{k'}\trp -
\vx_k \vx_j\trp - \vx_i \vx_{k'}\trp + \vx_i \vx_j\trp \rangle_{q(\vx)}.
\end{align*}
Because $\mathbf{L}$ is symmetric, $\mathbf{U}_{L}=\mathbf{V}_{L}$ in the SVD and 
$\mathbf{U}_L$ contains the eigenvectors of $\mathbf{L}$.
So $\mathbf{\tilde{L}}_{\epsilon}(p,q)=\mathbf{U}_{L}(p,n)\mathbf{U}_{L}(n,q)\frac{1}{\epsilon n}$
where we refer to the $n^{th}$ (last) eigenvector of $\mathbf{L}$. However, the
last eigenvector of $\mathbf{L}$ corresponding to the eigenvalue
0 has the same element in each coordinate i.e., $\mathbf{U}_{L}(:,n)=a\mathbf{1}_{n}$
for some constant $a\in\mathbb{R}$. This implies that
$\mathbf{\tilde{L}}_{\epsilon}(p,q)=aa\frac{1}{\epsilon n}$.
The elements of $\mathbf{\tilde{L}}_{\epsilon}$ have the same value, implying
$[ \b{\tilde{L}_\epsilon}(k,k') -  \b{\tilde{L}_\epsilon}(k,j) - \b{\tilde{L}_\epsilon}(i,k') +
\b{\tilde{L}_\epsilon}(i,j) ] = \frac{1}{\epsilon n}[aa - aa -aa + aa] =0$ and 
$\left\langle \boldsymbol{\Gamma}_{\epsilon,ij}\right\rangle =0$ for all $i,j$ blocks.  
We have 
 \begin{align*}
 l_1(\gamma) &=  - \frac{\gamma}{4} \mbox{Tr} ( \langle
 \hat{\b{Q}}  \tilde{\b{L}}_L \hat{\b{Q}}\trp \rangle_{q(\vx) } (d_y
 \b{\Sigma}_\b{C} + \vmu_{\b{C}}\trp \vmu_{\b{C}})). 
 \end{align*} 

 The second term $l_2(\gamma)$ is given by 
  \begin{align*} 
l_2(\gamma) &=  \gamma \mbox{Tr} (\vmu_{\b{C}}\trp \langle \b{H} \rangle_{q(\vx)} ), 
 \end{align*} 
 and the third term   $l_3(\gamma)$ is rewritten as 
   \begin{align*} 
l_3(\gamma) &=  -  \gamma \mbox{Tr}( \b{L} \b{Y}\trp \b{Y}). 
 \end{align*}
Finally, the last term is simplified as 
\begin{align*}
\frac{d_y}{2} \log | \epsilon \b{1}_n \b{1}_n\trp + 2 \gamma \b{L}| 
&= \frac{d_y}{2} \left(\sum_{i=1}^{n-1} \log \b{D}_L(i,i) + \log(n \epsilon) + (n-1) \log(2 \gamma) \right).
 \end{align*} Hence, 
 \begin{align*}
 l_4(\gamma) &= \frac{d_y}{2}(n-1) \log(2\gamma). 
 \end{align*}
The update for $\gamma$ is thus given by 
\begin{align*}
   \gamma &= \arg\max_\gamma l(\gamma)  
   = \arg\max_\gamma l_1(\gamma) + l_2(\gamma) +l_3(\gamma) + l_4(\gamma)   \\
   &= -\frac{d_y(n-1)/2}{
  - \frac{1}{4} \mbox{Tr} ( \langle
 \hat{\b{Q}}  \tilde{\b{L}}_L \hat{\b{Q}}\trp \rangle_{q(\vx) } (d_y
 \b{\Sigma}_\b{C} + \vmu_{\b{C}}\trp \vmu_{\b{C}}))
 +\mbox{Tr} (\vmu_{\b{C}}\trp \langle \b{H} \rangle_{q(\vx)} )
 -\mbox{Tr}( \b{L} \b{Y}\trp \b{Y})
 }.
\end{align*}

\subsubsection*{Update for $\alpha$}
We  update $\alpha$ by maximizing \eqref{free_energy} which is equivalent to maximizing the 
following expression.
\begin{align} 
-D_{KL}(q(\vx)||p(\vx|\b{G}, \hat{\vtheta}))
&= \Em_{q(\vx) q(\mathbf{C})}[\log p(\vx|\b{G}, \vtheta) - \log q(\vx) ] \\
&= - \int d\vx \; \Nrm(\vx|\vmu_\vx, \b{\Sigma}_\vx) \log \frac{\Nrm(\vx|\vmu_\vx, \b{\Sigma}_\vx)}{\Nrm(\vx|\b{0}, \b{\Pi})}, \nonumber \\
&= \frac{1}{2} \log |\b{\Sigma}_\vx \b{\Pi}^{-1} | - \frac{1}{2} \mbox{Tr} \left[ \b{\Pi}^{-1} \b{\Sigma}_\vx - \b{I}_{nd_x} \right] - \frac{1}{2} \vmu_\vx\trp \b{\Pi}^{-1} \vmu_\vx, \\
&= \frac{1}{2} \log |\b{\Sigma}_\vx| + \frac{1}{2} \log| \alpha \b{I} +
\b{\Omega}^{-1} | - \frac{\alpha}{2} \mbox{Tr} \left[ \b{\Sigma}_\vx \right]-
\frac{1}{2} \mbox{Tr} \left[\b{\Omega}^{-1} \b{\Sigma}_\vx \right]  \nonumber \\ 
& + \frac{nd_x}{2}
-\frac{\alpha}{2} \vmu_\vx\trp  \vmu_\vx - \frac{1}{2} \vmu_\vx\trp
\b{\Omega}^{-1} \vmu_\vx \\
&:= f_\alpha(\alpha)
\end{align}
The stationarity condition of $\alpha$ is given by
\begin{align}
 \frac{\partial}{\partial \alpha}\Em_{q(\vx) q(\b{C}) }[\log p(\vx|\b{G}, \vtheta) - \log q(\vx) ] 
=\frac{1}{2} \mbox{Tr}( ( \alpha \b{I} + \b{\Omega}^{-1})^{-1} ) - \frac{1}{2}
\mbox{Tr} \left[ \b{\Sigma}_\vx \right] - \frac{1}{2} \vmu_\vx\trp  \vmu_\vx =0,
\end{align} 
which is not closed-form and requires finding the root of the equation. 

For updating $\alpha$, we will find $\alpha = \argmax_\alpha f_\alpha(\alpha)$:
\begin{align}
    \alpha &= \argmax_\alpha  \log| \alpha I +
\b{\Omega}^{-1} | - \alpha\mbox{Tr} \left[ \b{\Sigma}_\vx \right]
- \alpha \vmu_\vx\trp  \vmu_\vx.
\label{eq:alpha_mstep_obj}
\end{align}

Assume $\b{\Omega}^{-1} = E_\Omega V_\Omega E_\Omega^\top$ by eigen-decomposition and 
$V_\Omega = \diag\left(v_{11}, .\ldots, v_{n d_x, n d_x} \right)$. The main difficult 
in optimizing $\alpha$ comes from the first term.
\begin{align}
    \log |\alpha I + \b{\Omega}^{-1}|  &\overset{(a)}{=} \log |\alpha E_\Omega
    E_\Omega^\top + E_\Omega V_\Omega E_\Omega^\top|  \\
    & = \log |E_\Omega (\alpha I +V_\Omega) E_\Omega^\top | \\ 
    & \overset{(b)}{=} \log |\alpha I +V_\Omega|  
    = \sum_{i=1}^{n d_x} \log(\alpha + v_{ii})  \\
    &\overset{(c)}{=} d_x\sum_{j=1}^{n}  \log(\alpha + 2\omega_i)
\end{align}
where at $(a)$ we use the fact that $E_\Omega$ is orthogonal. At $(b)$, 
the determinant of a product is the product of the determinants, and that the determinant 
of an orthogonal matrix is 1. Assume that $L = E_L V_L E_L^\top$ by eigen-decomposition 
and $V_L = \diag\left( \left\{ \omega_i \right\}_{i=1}^n \right)$. 
Recall that $\b{\Omega}^{-1} = 2 \b{L} \otimes \b{I}_{d_x}$.
By Theorem~\ref{thm:eigen_kron}, 
$v_{ii} = 2\omega_i$ and $2\omega_i$ appears $d_x$ times for each $i=1,\ldots, n$.
This explains the $d_x$ factor in $(c)$.

In the implementation, we use fminbnd in Matlab to optimize the negative of 
\eqref{alpha_mstep_obj} to get an update for $\alpha$. The eigen-decomposition of 
$L$ (not $\Omega^{-1}$ which is bigger) is needed only once in the beginning.
We only need the eigenvalues of $L$, not the eigenvectors.

\subsubsection*{KL divergence of $\b{C}$}
\begin{align*}
 & - D_{KL}(q(\b{C})||p(\b{C}|\b{G}))  \\ 
=& \Em_{q(\vx) q(\mathbf{C})}[\log p(\mathbf{C}|\b{G}, \vtheta) - \log q(\vc)]  \\
=& - \int d\vc \; \Nrm(\vc|\vmu_\vc, \b{\Sigma}_\vc) \log \frac{\Nrm(\vc|\vmu_\vc, \b{\Sigma}_\vc)}{ \Nrm(0,   ((\epsilon \b{J}  \b{J} \trp+ \b{\Omega}^{-1}) \otimes \b{I})^{-1})}, \nonumber \\
=& \frac{1}{2} \log |\b{\Sigma}_\vc((\epsilon \b{J}  \b{J} \trp+ \b{\Omega}^{-1}) \otimes \b{I}) | - \frac{1}{2} \mbox{Tr} \left[ \b{\Sigma}_\vc ((\epsilon \b{J}  \b{J} \trp+ \b{\Omega}^{-1}) \otimes \b{I}) - \b{I} \right] - \frac{1}{2} \vmu_\vc\trp ( (\epsilon \b{J}  \b{J} \trp+ \b{\Omega}^{-1}) \otimes \b{I} ) \vmu_\vc, \nonumber \\
=& \frac{1}{2} \log |(\b\Sigma_\b{C} (\epsilon \b{J}\b{J}\trp + \b{\Omega}^{-1})) \otimes \b{I} | - \frac{1}{2} \mbox{Tr} \left[(\b\Sigma_\b{C} (\epsilon \b{J}\b{J}\trp + \b{\Omega}^{-1})) \otimes \b{I}    - \b{I} \right] -  \frac{1}{2} \mbox{Tr}( (\epsilon \b{J}\b{J}\trp + \b{\Omega}^{-1}) \vmu_\b{C}\trp \vmu_\b{C}), \nonumber \\
=&  \frac{d_y}{2} \log |\b\Sigma_\b{C} (\epsilon \b{J} \b{J}\trp + \b{\Omega}^{-1})|  - \frac{d_y}{2}  \mbox{Tr} [ \b\Sigma_\b{C}(\epsilon \b{J}\b{J}\trp + \b{\Omega}^{-1})] + \frac{1}{2} nd_xd_y  -  \frac{1}{2} \mbox{Tr}((\epsilon \b{J}\b{J}\trp + \b{\Omega}^{-1})  \vmu_\b{C}\trp \vmu_\b{C}). \nonumber
%
\end{align*}

\section{Connection to GP-LVM}  
To see how our model is related to GP-LVM, we integrate out $\b{C}$ from the likelihood:
\begin{eqnarray*}
p(\vy|\vx, \b{G}, \vtheta) &=& \int p(\vy| \vc, \vx,  \vtheta) p(\vc|\b{G}) d\vc, \\
&\propto& \int \exp \left[- \frac{1}{2} (\vc\trp (\b{\Gamma} \otimes \b{I}) \; \vc - 2 \vc\trp \mbox{vec}(\b{V}^{-1} \b{H})) - \frac{1}{2} \vy\trp \Sigma_\vy^{-1} \vy - \frac{1}{2} \vc\trp ((\epsilon \b{J} \b{J}\trp + \b{\Omega}^{-1}) \otimes \b{I}) \; \vc  \right] d\vc, \\
&\propto&  \int \exp \left[- \frac{1}{2} (\vc\trp ((\b{\Gamma} + \epsilon \b{J} \b{J}\trp + \b{\Omega}^{-1}) \otimes \b{I} ) \; \vc - 2 \vc\trp \mbox{vec}(\b{V}^{-1} \b{H}))   \right] d\vc - \frac{1}{2} \vy\trp \Sigma_\vy^{-1} \vy, \\
&\propto& \exp\left[\frac{1}{2}  \mbox{vec}(\b{V}^{-1} \b{H})\trp ((\b{\Gamma} + \epsilon \b{J} \b{J}\trp + \b{\Omega}^{-1}) \otimes \b{I} )^{-T} \; \mbox{vec}(\b{V}^{-1} \b{H}) - \frac{1}{2} \vy\trp \Sigma_\vy^{-1} \vy \right], \nonumber 
\end{eqnarray*} where the last line comes from the fact : $\int \exp\left[-\frac{1}{2} \vc\trp \b{M}\vc + \vc\trp \vm \right] d \vc \propto \exp\left[\frac{1}{2} \vm\trp \b{M}^{-T} \vm \right]$. 

The term $\mbox{vec}(\b{V}^{-1} \b{H}) $ is linear in $\vy$ where 
\begin{eqnarray*}
\b{H}_i &=& \sum_{j=1}^n \eta_{ij} (\vy_j - \vy_i) (\vx_j - \vx_i)\trp, \\
&=& \sum_{j=1}^n \vy_j \eta_{ij} (\vx_j - \vx_i)\trp -  \vy_i  \sum_{j=1}^n \eta_{ij} (\vx_j - \vx_i)\trp, \\
&=& \tilde{\b{Y}} \vu_i + \vy_i \vv_i, 
\end{eqnarray*} where the vectors $\vu_i$ and $\vv_i$ are defined by 
\begin{eqnarray*}
\tilde{\b{Y}} &=& [\vy_1 \;  \cdots \; \vy_n], \\
\vu_i &=& \begin{bmatrix}
       \eta_{i1} (\vx_1- \vx_i)\trp \\[0.3em]
        \vdots \\[0.3em]
        \eta_{in} (\vx_n- \vx_i)\trp
\end{bmatrix}, \quad 
\vv_i = - \sum_{j=1}^n \eta_{ij} (\vx_j - \vx_i)\trp. 
\end{eqnarray*} Using these notations, we can write $H$ as
\begin{eqnarray*}
\b{H} &=& [ \b{H}_1, \cdots, \b{H}_n], \\
&=& \tilde{\b{Y}} \b{W}, 
\end{eqnarray*} where 
\begin{eqnarray*}
\b{W} &=& \b{U}_u + \b{V}_v, \\ 
\b{U}_u &=& [\vu_1, \cdots, \vu_n], \\
\b{V}_v &=& \begin{bmatrix}
       \vv_1 & 0 & \cdots & 0  \\[0.3em]
       0 & \vv_2 & \cdots & 0  \\[0.3em]
        \vdots &0  & \vdots & 0 \\[0.3em]
       0 & \cdots & 0 & \vv_n  
\end{bmatrix}.
\end{eqnarray*} So, we can explicitly write down $\mbox{vec}(\b{V}^{-1} \b{H}) $  as 
\begin{eqnarray*}
\mbox{vec}(\b{V}^{-1} \b{H}) &=& \mbox{vec}(\b{V}^{-1} \tilde{\b{Y}} \b{W}), \\
&=& (\b{W}\trp \otimes \b{V}^{-1})\mbox{vec}(\tilde{\b{Y}}), \\
&=&  (\b{W}\trp \otimes \b{V}^{-1}) \vy. 
\end{eqnarray*}

Using all these, we can rewrite the likelihood as 
\begin{eqnarray*}
p(\vy|\vx, \b{G}, \vtheta) &\propto&  \exp\left[ - \frac{1}{2} \vy\trp \; \b{K}^{-1}_{LL} \; \vy \right], 
\end{eqnarray*} where the precision matrix is given by 
\begin{eqnarray*}
\b{K}^{-1}_{LL}&=& \b{\Sigma}_\vy^{-1} - (\b{W}\trp \otimes \b{V}^{-1})\trp \b{\Lambda} (\b{W}\trp \otimes \b{V}^{-1}), \\
\b{\Lambda} &=&((\b{\Gamma} + \epsilon \b{J} \b{J}\trp + \b{\Omega}^{-1}) \otimes \b{I} )^{-T} . 
\end{eqnarray*}

\section{Useful results}  
In this section, we summarize theorems and matrix identities useful for deriving 
update equations of LL-LVM.  The notation in this section is independent of the rest.

\begin{theorem}
    \label{thm:eigen_kron}
    Let $A \in \mathbb{R}^{n\times n}$ have eigenvalues $\lambda_i$, and let $B \in \mathbb{R}^{m \times m}$
    have eigenvalues $\mu_j$. Then the $mn$ eigenvlaues of $A \otimes B$ are 
    \begin{equation*}
        \lambda_1\mu_1, \ldots, \lambda_1\mu_m, \lambda_2 \mu_1,\ldots, \lambda_2\mu_m, \ldots, \lambda_n \mu_m.
    \end{equation*}
\end{theorem}

\begin{theorem}
    A graph Laplacian $L \in \mathbb{R}^{n \times n}$ is positive semi-definite.
    That is, its eigenvalues are non-negative.
\end{theorem}

\subsection{Matrix identities}
\begin{align}
x^{\top}(A\circ B)y=\trace(\diag(x)A\diag(y)B^{\top}) 
\end{align}

From section 8.1.1 of the matrix cookbook \cite{Petersen2012},
\begin{align}
\int\exp\left[-\frac{1}{2}x^{\top}Ax+c^{\top}x\right]\,\mathrm{dx}=\sqrt{\det(2\pi
    A^{-1})}\exp\left[\frac{1}{2}c^{\top}A^{-\top}c\right].  
\end{align}
    
\begin{lemma}
    If
    $X=\left(x_{1}|\cdots|x_{n}\right)$ and $C=\left(c_{1}|\ldots|c_{n}\right)$,
    then
    \begin{equation*}
    \int\exp\left[-\frac{1}{2}\trace(X^{\top}AX)+\trace(C^{\top}X)\right]\:\mathrm{d}X=\det(2\pi
    A^{-1})^{n/2}\exp\left[\frac{1}{2}\trace(C^{\top}A^{-1}C)\right].
    \end{equation*}
\end{lemma}
 
Woodbury matrix identity
\begin{align}
    \label{eq:woodbury}
    (A + UCV)^{-1} = A^{-1} - A^{-1} U (C^{-1} + VA^{-1}U)^{-1} VA^{-1}.
\end{align}

\newpage

\end{document}